\documentclass{bmvc2k}
\usepackage{times}
\usepackage{epsfig}
\usepackage{graphicx,graphics}
\usepackage{amsmath}
\usepackage{amssymb}
\usepackage{float}
\usepackage{bbm}
\usepackage{multirow}
\usepackage{graphicx}
\usepackage{pifont}
\newcommand{\cmark}{\ding{51}}
\newcommand{\xmark}{\ding{55}}
\usepackage{wrapfig}
\usepackage{amsmath}

\usepackage[font=small,skip=0pt]{caption}

\title{BEA: Revisiting anchor-based object detection DNN using Budding Ensemble Architecture}

\addauthor{Syed Sha Qutub}{syed.qutub@intel.com}{1,2}
\addauthor{Neslihan Kose}{neslihan.kose.cihangir@intel.com}{1}
\addauthor{Rafael Rosales}{rafael.rosales@intel.com}{1}
\addauthor{Michael Paulitsch}{michael.paulitsch@intel.com}{1}
\addauthor{Korbinian Hagn}{Korbinian.Hagn@intel.com}{1}
\addauthor{Florian Geissler}{florian.geissler@intel.com}{1}
\addauthor{Yang Peng}{yang.r.peng@intel.com}{1}
\addauthor{Gereon Hinz}{gereon.hinz@tum.de}{2}
\addauthor{Alois Knoll}{knoll@in.tum.de}{2}

\addinstitution{
 Intel Labs \\
 Munich, Germany
}
\addinstitution{
 Technical University of Munich \\
 Munich, Germany
}

\runninghead{Syed \etal}{BEA}

\def\etal{\emph{et al}\bmvaOneDot}

\begin{document}

\maketitle

\begin{abstract}
This paper introduces the Budding Ensemble Architecture (BEA), a novel reduced ensemble architecture for anchor-based object detection models. Object detection models are crucial in vision-based tasks, particularly in autonomous systems. They should provide precise bounding box detections while also calibrating their predicted confidence scores, leading to higher-quality uncertainty estimates. However, current models may make erroneous decisions due to false positives receiving high scores or true positives being discarded due to low scores. BEA aims to address these issues.
The proposed loss functions in BEA improve the confidence score calibration and lower the uncertainty error, which results in a better distinction of true and false positives and, eventually, higher accuracy of the object detection models. Both Base-YOLOv3 and SSD models were enhanced using the BEA method and its proposed loss functions. The BEA on Base-YOLOv3 trained on the KITTI dataset results in a 6\% and 3.7\% increase in mAP and AP50, respectively. Utilizing a well-balanced uncertainty estimation threshold to discard samples in real-time even leads to a 9.6\% higher AP50 than its base model. 
This is attributed to a 40\% increase in the area under the AP50-based retention curve used to measure the quality of calibration of confidence scores. Furthermore, BEA-YOLOV3 trained on KITTI provides superior out-of-distribution detection on Citypersons, BDD100K, and COCO datasets compared to the ensembles and vanilla models of YOLOv3 and Gaussian-YOLOv3.
\end{abstract}

\section{Introduction}
\label{sec:intro}
\begin{figure}[h]
    \bmvaHangBox{\fbox{\includegraphics[width=0.48\textwidth, scale=0.5]{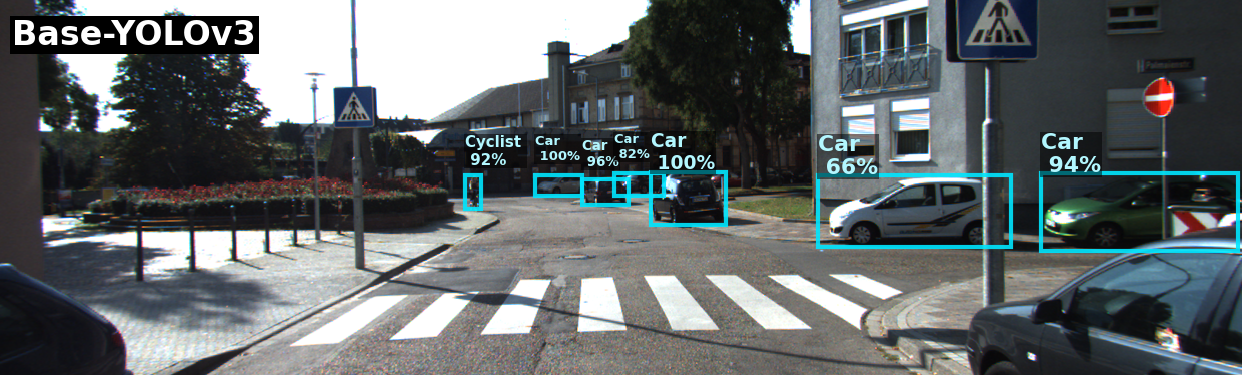}}}
    \bmvaHangBox{\fbox{\includegraphics[width=0.48\textwidth, scale=0.5]{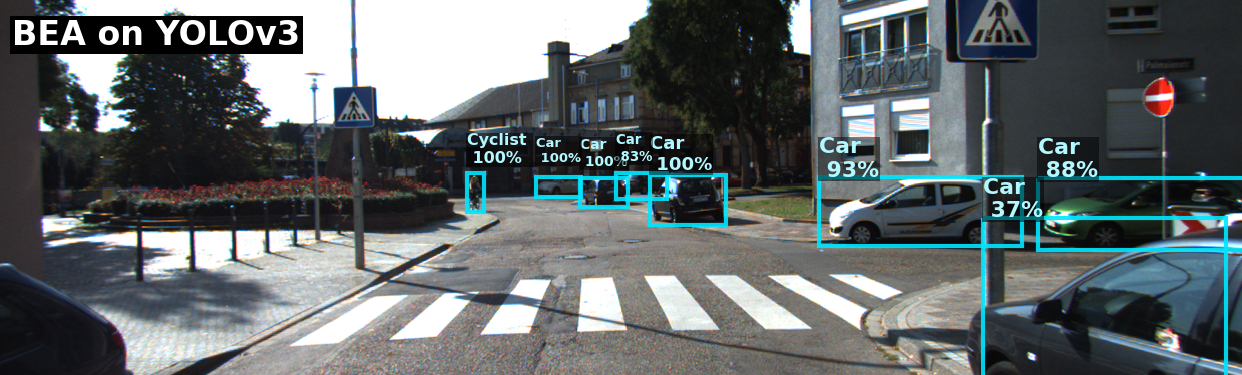}}}
\caption{The figure on the left demonstrates uncalibrated confidence scores of Base-YOLOv3, where the white car has a 66\% confidence score despite negligible occlusion, and the green car on the right edge has a 94\% confidence score despite occlusion. In contrast, the right figure shows that BEA-YOLOv3 detects a car on the bottom right edge, which Base-YOLOv3 misses, and has superior confidence score calibration, indicating the effectiveness of our approach over baseline (Base-YOLOv3).}
\label{fig:BEA_image1}
\end{figure}

Object detection deep neural networks (DNN) are a critical technology in numerous fields, such as medical imaging, robotics, and autonomous vehicles.
The current state-of-the-art in object detection models can be widely classified into single-stage ( YOLOv3 \cite{redmon2018yolov3}, Retina-Net \cite{lin2017focal}, SSD \cite{liu2016ssd}) and two-stage detectors (Faster r-cnn \cite{ren2015faster}, Fast-r-cnn \cite{girshick2015fast}, Mask-RCNN \cite{he2017mask}). 
Two-stage detectors are the current state-of-the-art for object detection, as they propose regions of interest before predicting object class and bounding boxes. Single-stage detectors, which directly predict object class and bounding boxes from anchor or prior boxes, are faster but less accurate. Anchor-based single-stage detectors are suitable for real-time applications due to their lower computational overhead. To improve the accuracy of object detection models, researchers have proposed ensemble modelling \cite{miller2019evaluating} and post-hoc calibration \cite{guo2017calibration} techniques.
There are different ensemble-based approaches, and they generally perform better than a single standalone model, mainly for the following reasons. It captures uncertainty when models are trained with subsets of extensive data. They are more robust to over-fitting as the predictions are averaged, indirectly improving the generalization towards out-of-distribution data. Similarly, the post-hoc calibration techniques are the methods to re-calibrate the predicted probabilities of a model post-training. They are better than no calibration of a model as the confidence scores cannot be directly trusted, but it deals with its inherent issues and is limited by the model's training method. It is also highly biased towards the dataset it is trained with. In \cite{Ovadia2019}, the authors claim that post-hoc methods fail to provide well-calibrated uncertainty estimates under distributional shifts in the real world.

This work aims to enhance the quality of the prediction confidence score through calibration (as shown in Figure ~\ref{fig:BEA_image1}). Our approach builds upon the work of Kornblith \etal \cite{kornblith2019similarity}, who demonstrated that wider networks with similar architecture tend to learn more similar feature representations. By measuring multiple similarity scores, they found that the similarity of early layers in these networks plateaus at fewer channels than later layers and that early layers learn comparable representations across diverse datasets. By leveraging these insights, we propose a more efficient and effective reduced ensemble architecture called the \textbf{Budding Ensemble Architecture (BEA)}, which includes a common backbone and only two replicated detectors. 
The name BEA reflects that its two detectors branch out from the backbone. We extensively evaluate our proposed method and demonstrate its superior performance to the current state-of-the-art through a series of experiments on the KITTI dataset \cite{Geiger2012CVPR}. The BEA is equipped with a novel function: \textbf{Tandem} loss to enhance calibration and capture out-of-distribution uncertainty in the predictions. The proposed approach results in well-calibrated predictions leading to better uncertainty estimates and aiding in detecting out-of-distribution (OOD) images.
Furthermore, we show the quality of the predictions using average precision metric and the quality of our uncertainty measure using uncertainty error (UE) ~\cite{miller2019evaluating} and Retention curves ~\cite{malinin2021shifts}.

In summary, this work makes the following contributions:
\begin{enumerate}

\item This study presents the Budding Ensemble Architecture (BEA), a novel architecture that outperforms state-of-the-art models in terms of accuracy. The experimental results demonstrate that the BEA enhances confidence score calibration more than the base and ensemble models.
\item We introduce AP50-based retention curves to measure the calibration quality for object detection models.
\item The BEA demonstrates promising results in capturing more accurate detection of out-of-distribution images compared to their respective vanilla and ensemble models.
\end{enumerate}

\section{Related Work}
Object detection is a complex, high-dimensional task that involves detecting objects within an image and accurately determining their location and size using confidence scores and regression techniques. This makes uncertainty quantification challenging, as many potential error sources must be considered.
There are two main approaches to uncertainty estimation in neural network predictions: Bayesian \cite{welling2011bayesian, blundell2015weight, gal2016dropout, zhang2020csgmcmc, pmlr2020BNN} and non-Bayesian \cite{lakshminarayanan2017simple, liu2020simple, pmlr2020}. Bayesian approaches use probabilistic models to address uncertainty but are computationally demanding, while non-Bayesian methods use heuristics or techniques to estimate uncertainty, resulting in lower accuracy. Proposed methods for well-calibrated uncertainty estimation primarily focus on classification tasks or are based on post-hoc calibration.
In \cite{AvUC20}, the authors target classification tasks and introduce differentiable accuracy versus uncertainty calibration (AvUC) loss to provide well-calibrated uncertainties and improved accuracy. In \cite{kose2023reliable}, the authors propose a loss-calibrated inference method for time-series-based prediction and regression tasks following the insights from \cite{AvUC20}. 
Post-hoc methods \cite{Kuleshov18, kumar2019verified, UAI21} also mainly target classification or small-scale regression problems. These methods are based on recalibrating a pre-trained model using a sample of independent and identically distributed (i.i.d.) data.  In \cite{Kuleshov18}, the authors propose a post-hoc method to calibrate the output of 1D regression algorithms based on Platt scaling \cite{Platt99probabilisticoutputs}. The authors in \cite{schubert2021metadetect} explored the calibration properties of object detection networks and examined the influence of position and shape in object detection on the calibration properties. Gaussian-YOLOv3~\cite{choi2019gaussian} introduces Gaussian modelling of bounding box predictions, where variances are utilized as prediction uncertainty. However, evaluating its performance in terms of confidence score calibration is inadequate. To the best of our knowledge, there are limited numbers of existing calibration solutions for object detection as they bring additional challenges when designing these methods.
Ensemble deep neural networks (DNNs) improve DNN model performance and robustness. One way to build ensemble DNNs is to train multiple models with different initializations, architectures, or input data and combine their predictions. This paper proposes a novel approach to enhance the reliability of uncertainty quantification in anchor-based object detection networks by introducing redundant layers after the backbone feature extractor. Inspired by the Siamese architecture \cite{bromley1993signature, chicco2021siamese}, the proposed approach aims to maximize true positives while minimizing false positives using a differentiable loss function and architecture. This leads to highly calibrated models with improved confidence scores and accurate bounding boxes for detected objects, improving reliability and performance. 
Detecting out-of-distribution (OOD) images is a crucial task in various machine learning and computer vision applications to ensure model robustness and safety.

Existing OOD detection literature mainly focuses on classification networks \cite{yang2021generalized}, employing post-hoc methods like temperature scaling \cite{liang2017enhancing}, confidence enhancement techniques \cite{devries2018learning}, or training with OOD data \cite{dhamija2018reducing}. 
For instance, a multitude of out-of-distribution (OOD) detectors has been developed using generative models such as Variational Autoencoders (VAEs) \cite{xiao2020likelihood, ran2022detecting} and Generative Adversarial Networks (GANs) \cite{ryu2018out}. These detectors typically rely on indicators like reconstruction error or assessing the likelihood of a sample being generated by the model. However, these solutions often come with additional computational overhead, primarily dedicated to OOD detection rather than conventional object detection.
In contrast, our proposed BEA-based models offer a dual functionality and minimal overhead. They perform standard object detection seamlessly while concurrently delivering state-of-the-art OOD detection capabilities, thus offering a more efficient and integrated solution.
\vspace{-5mm}
\section{Budding Ensemble Architecture}
\label{sec:bea_on_yolov3}

This section introduces the Budding Ensemble Architecture (BEA), a novel architecture comprising a shared backbone and only two duplicated detectors that is responsible for enhancing the efficiency and effectiveness of this reduced ensemble model compared to traditional object detection ensembles. The BEA not only calibrates prediction confidence scores but also enables the detection of out-of-distribution (OOD) samples. Further, the application of BEA is explained in detail using YOLOv3 model as an example.


\begin{figure*}
\centering
  \begin{minipage}[b]{0.39\linewidth}
    \centering
    \begin{subfigure}{}
        \fbox{\includegraphics[width=0.65\linewidth]{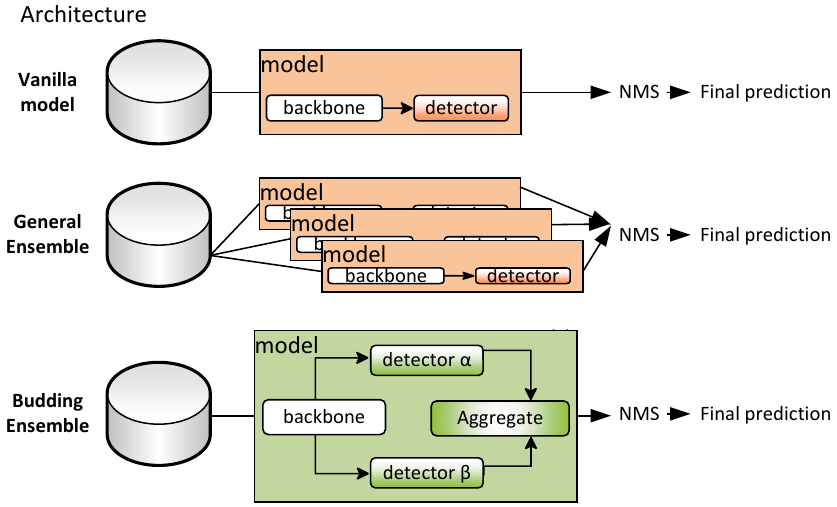}}
      \caption{Existing architectures vs Budding Ensemble architecture}
      \label{fig:simpleBEA}
    \end{subfigure}
    \begin{subfigure}{}
      \fbox{\includegraphics[width=0.65\linewidth]{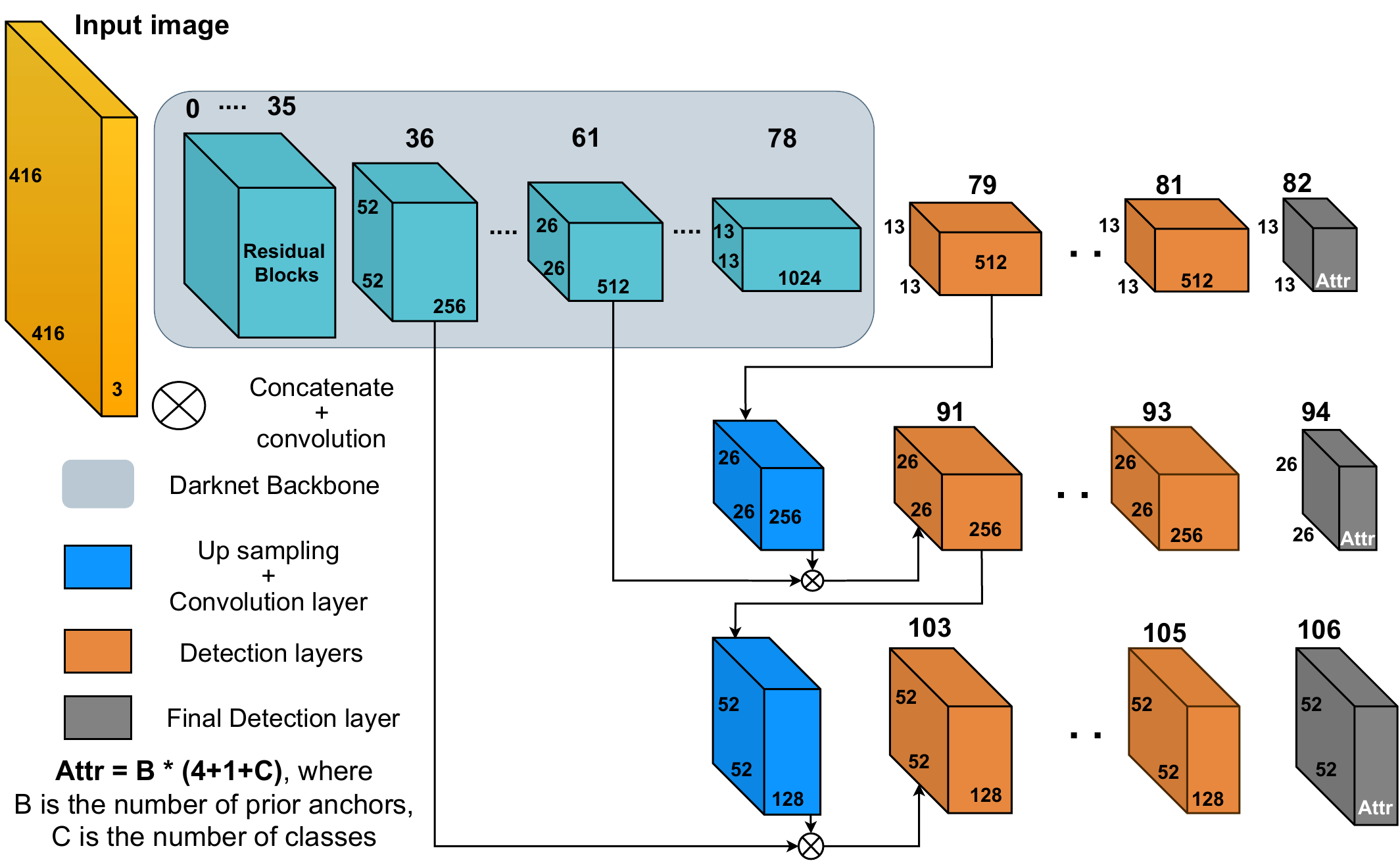}}
      \caption{Base model: YOLOv3 architecture}
      \label{fig:vanilla_YoloV3}
    \end{subfigure}
  \end{minipage}
  \begin{minipage}[b]{0.59\linewidth}
    \centering
  \fbox{\includegraphics[width=0.75\linewidth]{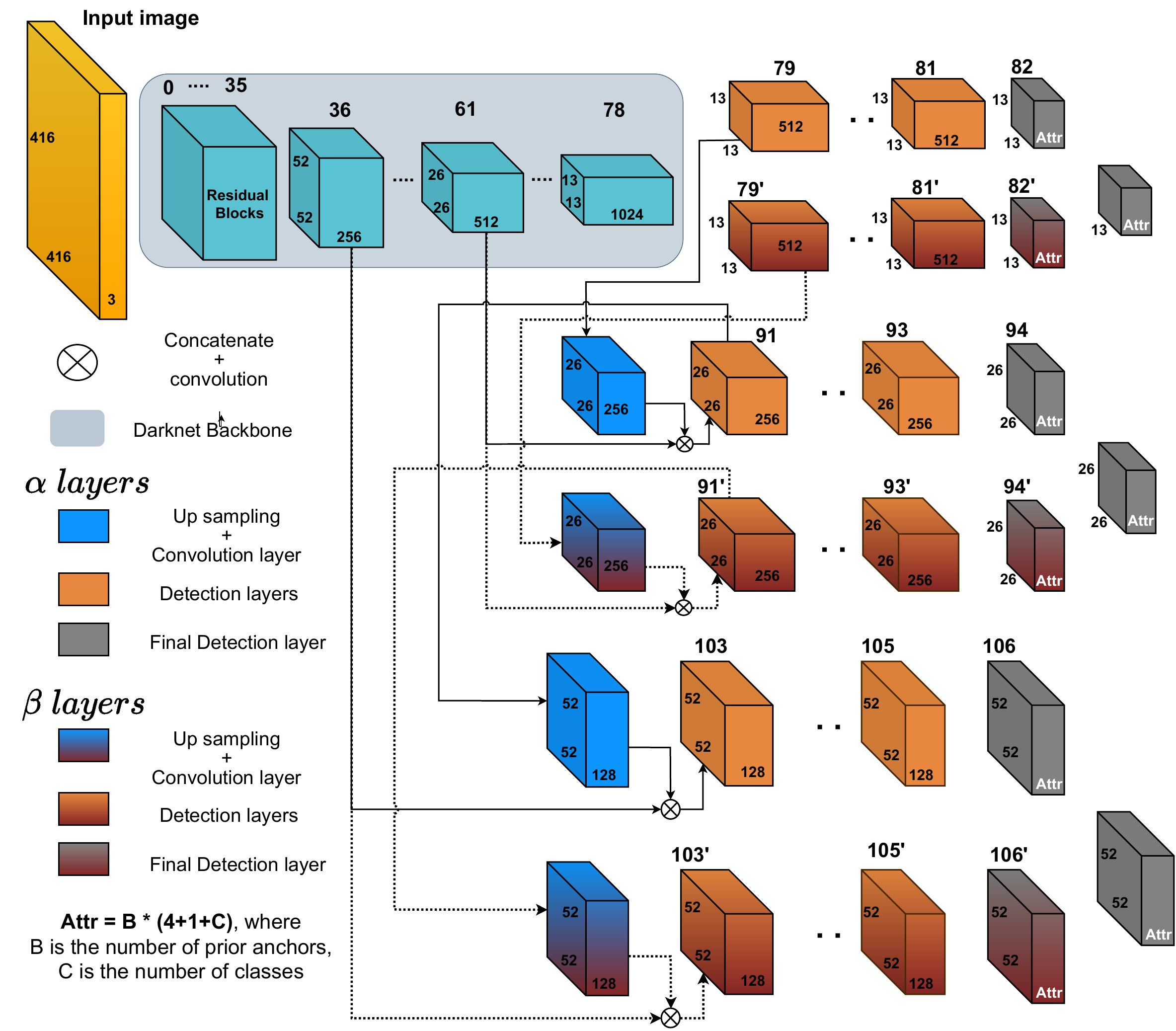}}
  \caption{Budding Ensemble Architecture on YOLOv3}
  \label{fig:BEA_YoloV3}
  \end{minipage}
\end{figure*}

\subsection{BEA-YOLOv3}
\textbf{Recap on YOLOv3 (base model)}: In this model, three detectors are placed in the final layers, each dedicated to predicting objects of different scales, similar to the concept of feature pyramid networks (FPN \cite{lin2017feature}) illustrated in Fig.~\ref{fig:vanilla_YoloV3}. YOLOv3's final detector layers perceive the input image as \(\textit{\textbf{S x S}}\) grids, where each grid size $S$ is distinct for each detector layer. The final layer's prediction feature maps are tasked with object detection using three anchors (\textit{\textbf{B}}) with distinct aspect ratios on the corresponding grid of an input image. Therefore, there are three predictions per grid at each scale of the prediction maps in the final detection layers.

The YOLOv3 model is transformed into the BE Architecture shown in Figure ~\ref{fig:simpleBEA} and ~\ref{fig:BEA_YoloV3} by duplicating the detector layers. This results in six detectors, compared to the three in the base YOLOv3 model. We refer to the original and duplicate layers as \textbf{tandem layers}. 
Assuming that the initial setup of creating the tandem layers is accurate and combining the predictions from these layers is correct, it is reasonable to expect that the model's accuracy and confidence score calibration will enhance. This is because the model can analyze the intermediate features twice by sending the same representations to both detectors, which aids in capturing uncertainty.
However, during regular training of the vanilla form of BEA YOLOv3 architecture with its conventional loss functions (using $\mathcal{L_{\mathbf{conv}}}$), both the original detectors ($\mathcal{\alpha}$) and duplicate detectors ($\mathcal{\beta}$) end up learning similar representations, confidently agreeing on both correct and incorrect predictions.
To enhance the model's uncertainties and calibration, we aim for a strong disagreement on incorrect predictions and agreement on correct predictions. To achieve this, we propose two new loss functions: the Tandem-Quelling loss function ($\mathcal{L_{\mathbf{tq}}}$) (eq. \ref{eq:tandem_quelling_loss}) to encourage disagreement between negative predictions when no object is present in the grid, and the Tandem-Aiding loss function ($\mathcal{L_{\mathbf{ta}}}$) (eq. \ref{eq:tandem_aid_loss}) to promote agreement between positive predictions when an object is present at the individual output of both the regression (bounding box information x,y,w,h) and classification ($\boldsymbol{C}$). This setup also allows the model to exchange information when one of the tandem detectors is not confident about its prediction.
\begin{equation}
\label{eq:tandem_quelling_loss}
\scalebox{0.80}{$
 \mathfrak{L_{\mbox{tq}}(\hat{\phi})}=\sum_{i=1}^{S^2}\sum_{j=1}^{B} \mathbbm{1}_{\textit{ij}}^{\mbox{noobj}}  \frac{2} {\sqrt{\left(\hat{\phi}^{\mathcal{\alpha}}_i - \hat{\phi}^{\mathcal{\beta}}_i\right)^2}}, \quad \mathcal{L_{\mbox{tq}}}=\mathfrak{L_{\mbox{tq}}(\hat{\boldsymbol{x}})} + \mathfrak{L_{\mbox{tq}}(\hat{\boldsymbol{y}})} + \mathfrak{L_{\mbox{tq}}(\hat{\boldsymbol{w}})} + \mathfrak{L_{\mbox{tq}}(\hat{\boldsymbol{h}})} + \mathfrak{L_{\mbox{tq}}(\hat{\boldsymbol{C}})}
 $}
\end{equation}
\begin{equation}
\label{eq:tandem_aid_loss}
\scalebox{0.80}{$
    \mathfrak{L_{\mbox{ta}}(\hat{\phi})}=\sum_{i=1}^{S^2}\sum_{j=1}^{B} \mathbbm{1}_{\textit{ij}}^{\mbox{obj}}  \frac{\sqrt{\left(\hat{\phi}^{\mathcal{\alpha}}_i - \hat{\phi}^{\mathcal{\beta}}_i\right)^2}} {2}, \quad \mathcal{L_{\mbox{ta}}}=\mathfrak{L_{\mbox{ta}}(\hat{\boldsymbol{x}})} + \mathfrak{L_{\mbox{ta}}(\hat{\boldsymbol{y}})} + \mathfrak{L_{\mbox{ta}}(\hat{\boldsymbol{w}})} + \mathfrak{L_{\mbox{ta}}(\hat{\boldsymbol{h}})} + \mathfrak{L_{\mbox{ta}}(\hat{\boldsymbol{C}})}
$}
\end{equation}
\begin{equation}
\label{eq:tandem_loss}
\begin{aligned}
    \mathcal{L_{\mbox{tandem}}} &= \mathcal{L_{\mbox{tq}}} + \mathcal{L_{\mbox{ta}}} \\
\end{aligned}
\end{equation}
\begin{equation}
\label{eq:bea_loss}
\begin{aligned}
    \mathcal{L_{\mbox{bea}}} &= \mathcal{L_{\mbox{conv}}} + \mathcal{L_{\mbox{tandem}}} \\
\end{aligned}
\end{equation}

where $\mathbbm{1}_{\textit{ij}}$ in eq. ~\ref{eq:tandem_quelling_loss} and ~\ref{eq:tandem_aid_loss} denotes that the object in ground truth appears in $j_{th}$ anchor or the  bounding box predictor and grid $i$ is responsible for the particular prediction. $\hat{\phi}$ is a tuple consists of specific outputs from $\alpha$ and $\beta$ detector, $\hat{\phi}=(\hat{\phi}^\mathcal{\alpha}, \hat{\phi}^\mathcal{\beta})$.
Collectively the \textbf{Tandem loss functions} - $\mathcal{L_{\mathbf{tandem}}}$ (eq. \ref{eq:tandem_loss}) operate on the tandem detectors $\alpha$ and $\beta$ to increase the variance between their negative predictions. Similarly, they decrease the variances between the positive predictions.
\begin{figure}
\centering
\begin{tabular}{cccc}
\bmvaHangBox{\fbox{\includegraphics[width=2.6cm, height=1.9cm]{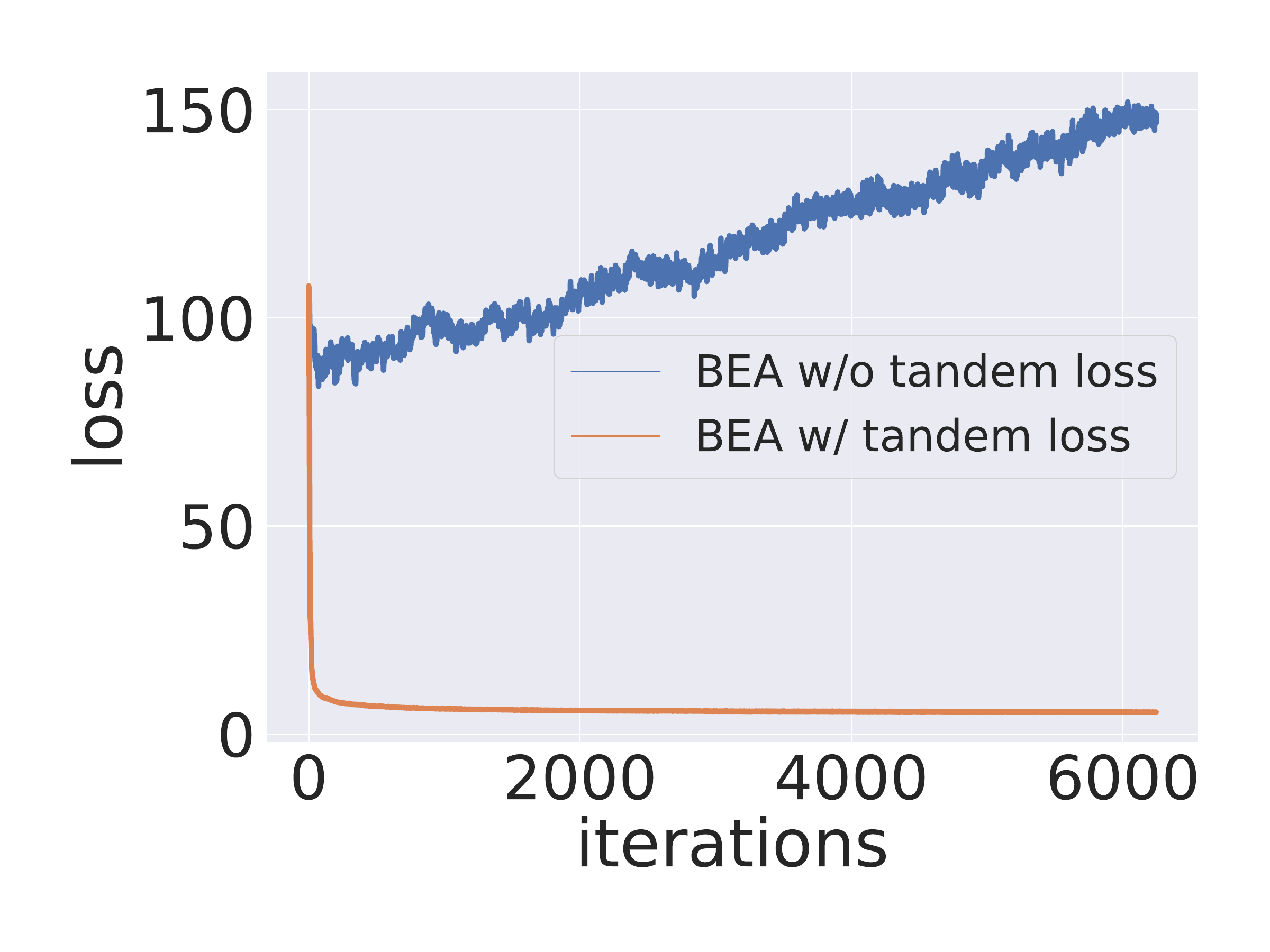}}}&
\bmvaHangBox{\fbox{\includegraphics[width=2.6cm, height=1.9cm]{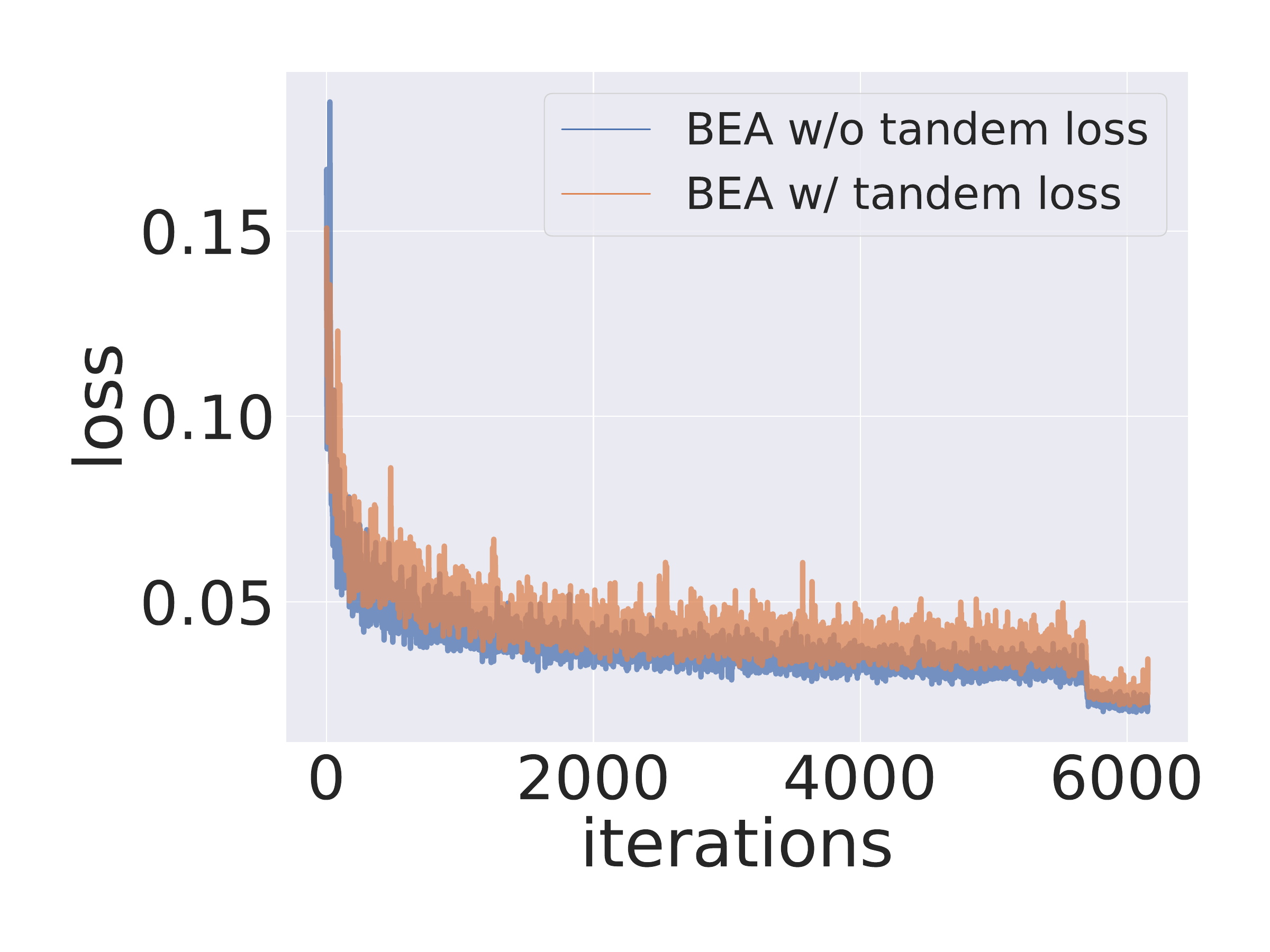}}}&
\bmvaHangBox{\fbox{\includegraphics[width=2.6cm, height=1.9cm]{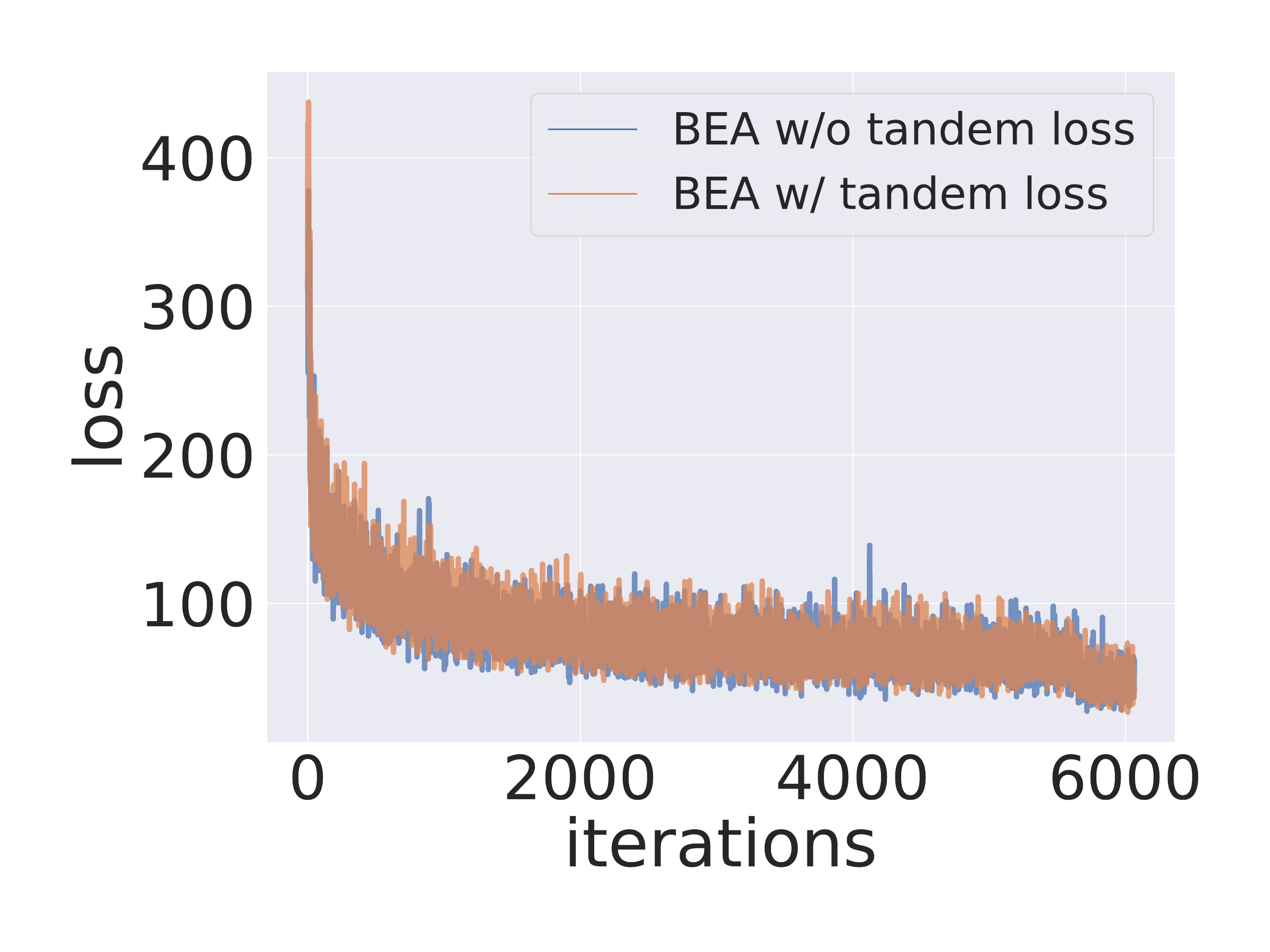}}}&
\bmvaHangBox{\fbox{\includegraphics[width=2.6cm, height=1.9cm]{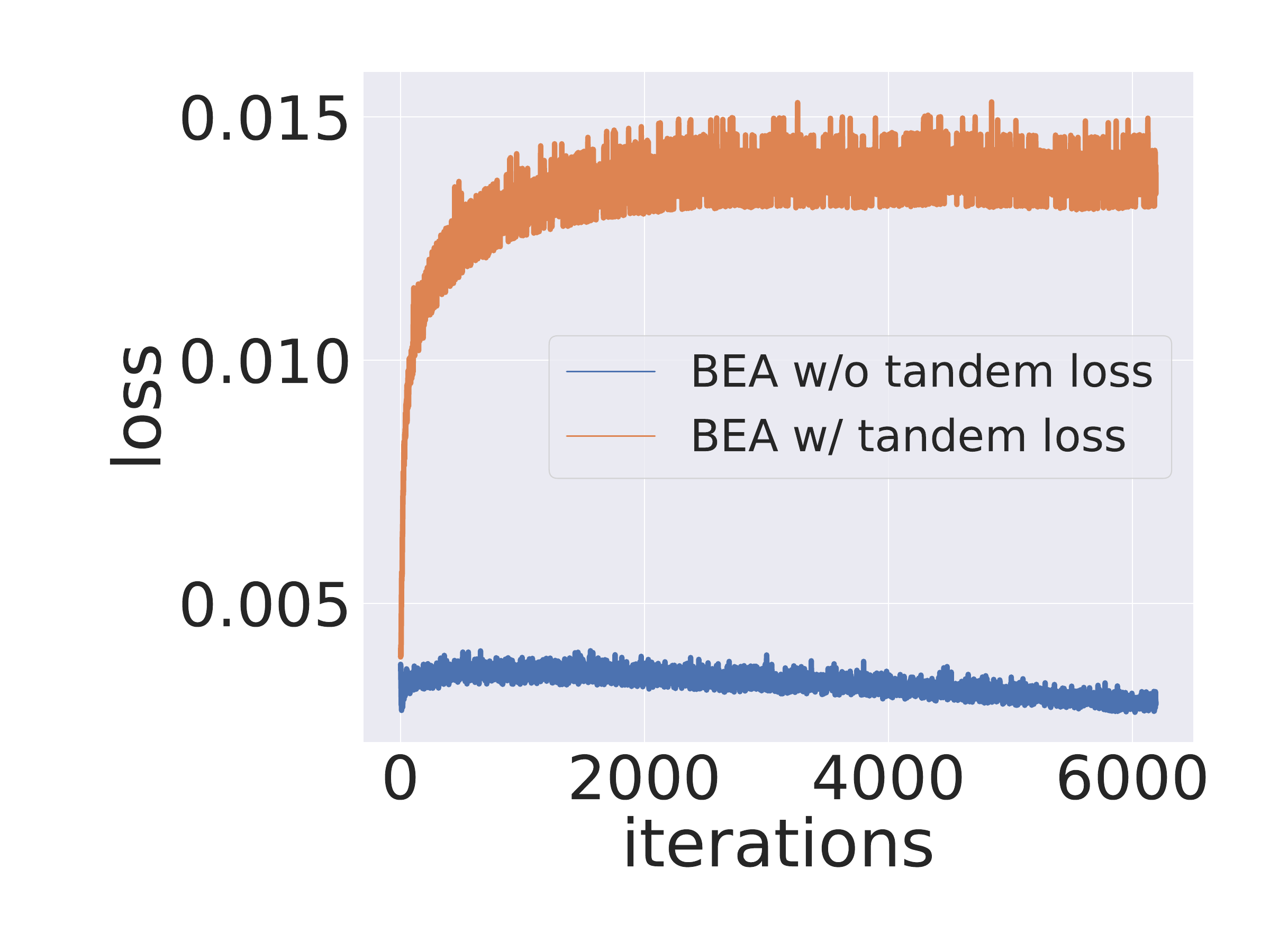}}}\\
(a)&(b)&(c)&(d)
\end{tabular}
\caption{Monitoring tandem loss functions ($\mathcal{L_{\mathbf{tandem}}}$) on BEA-YOLOv3. WH indicates predicted width and height offsets of the bounding box: (a) Monitoring tandem-quelling ($L_{\mbox{tq}}$) on WH; (b) Monitoring tandem-aiding ($L_{\mbox{ta}}$) on WH; (c) Original WH loss; (d) Normalised MSE between $\mathcal{\alpha}$ and $\mathcal{\beta}$ detectors}
\label{fig:monitoring_tandem_loss}
\label{fig:tandem_tq}
\label{fig:tandem_ta}
\label{fig:tandem_orig_wh}
\label{fig:tandem_norm_mse}
\end{figure}

\textbf{Aggregation of Tandem detectors}: During the inference stage, the tandem detectors are combined by averaging their bounding box coordinates, resulting in tighter boxes closer to the ground truth objects. However, only the maximum objectness and class scores from the tandem detectors are carried forward. Before discarding the tandem detectors, the $\mathcal{U}{ood}$ for OOD detection is captured at each anchor as explained in section ~\ref{evaluation_metrics_subsec}. Aggregated detector predictions are sent to a Non-maximum suppression (NMS) module, which uses modified bounding box coordinates to remove overlapping entities and lower confidence score predictions. The method involves maximum confidence score voting and averaging of bounding boxes.

\textbf{Uncertainty induced by Tandem loss functions:}
Fig. ~\ref{fig:monitoring_tandem_loss} shows how the individual factors of the loss function $\mathcal{L_{\mathbf{tandem}}}$ operate. The addition of $\mathcal{L_{\mathbf{tandem}}}$ to the vanilla BEA form increases the variance between negative predictions, causing the $\mathcal{L_{\mathbf{tq}}}$ loss to decrease, as seen in Fig. ~\ref{fig:tandem_tq}.
However, the introduction of $\mathcal{L_{\mathbf{tandem}}}$ does not affect the $\mathcal{L_{\mathbf{ta}}}$ loss (as shown in Fig. ~\ref{fig:tandem_ta}), since the original loss functions of YOLOv3 already operate on positive predictions individually for each detector. It is worth noting that the $\mathcal{L_{\mathbf{ta}}}$ loss plays a crucial role in reestablishing confidence in positive predictions with low variance between the tandem layers. The introduced loss function's effectiveness is validated by monitoring the mean squared error (mse) between the $\alpha$ and $\beta$ detectors (Fig. ~\ref{fig:tandem_norm_mse}). This approach helps suppress false positives during training and improves the model’s confidence score calibration, enhancing trust in the model.

\textbf{Tradeoff between $\mathcal{L_{\mbox{conv}}}$ and $\mathcal{L_{\mbox{tandem}}}$}: We conducted experiments by varying the weights assigned to $\mathcal{L_{\mbox{conv}}}$ and $\mathcal{L_{\mbox{tandem}}}$ to assess their influence on performance and uncertainty error. Assigning a weight greater than 1 to $\mathcal{L_{\mbox{tandem}}}$ did not lead to a significant reduction in uncertainty error or an enhancement in OOD detection. Instead, it had an adverse impact on prediction performance compared to the baseline.
To achieve a balanced improvement across predictions, reduction in uncertainty error, and enhanced OOD detection, it is imperative to allocate equal weight to both loss components.
\subsection{BEA on Gaussian-YOLOv3 and SSD}
To demonstrate the proposed method's effectiveness and versatility, we applied it to the Gaussian-YOLOv3 model \cite{choi2019gaussian} and SSD \cite{liu2016ssd}, as previously discussed in this section. Gaussian-YOLOv3 is a modified version of YOLOv3 where the bounding box coordinates (\^x, \^y, \^w, and \^h) are modelled as Gaussian outputs using negative log-likelihood (NLL) loss functions. Despite this alteration, the functionality of $\mathcal{L_{\mathbf{tandem}}}$ remains unchanged, as it operates on the $\mathcal{\alpha}$ and $\mathcal{\beta}$ detectors, whose predictions are generated by the Gaussian function. The BEA technique is also applied to SSD by replicating only the \textbf{Detections} layers, as illustrated in Figure 2 of the paper \cite{liu2016ssd}. Being an orthogonal approach, this demonstrates BEA's ability to be applied to different object detection models. The point of split where the detectors are duplicated is an interesting factor which can impact the efficiency of BEA. This is further discussed using SSD as an example in the supplementary material.

\section{Evaluation of BEA}
\label{evaluation_metrics}
\textbf{Experimental Setup}: We compare the performance of the proposed BEA architecture for object detection with different versions of YOLOv3 and SSD models, trained using a standardized approach with identical hyperparameters and an input image size of $416\times416$ using the mmdetection framework \cite{chen2019mmdetection}. YOLOv3 ensembles were trained for 300 epochs using the bagging method \cite{galar2011review} with pre-trained weights from COCO \cite{lin2014microsoft}. The evaluation was performed on the KITTI dataset \cite{Geiger2012CVPR} with seven usable classes out of nine, and the dataset was split using a ratio of 7.5:1:1.5 for training, validation, and testing to evaluate the model's generalization ability.
\subsection{Evaluation Metrics}
\label{evaluation_metrics_subsec}
In this section, we will discuss the evaluation metrics used to measure the effectiveness of applying BEA to YOLOv3 and SSD. The subsequent analysis delves into the experiments and their results in detail.

\textbf{Uncertainty Error (UE):} We use the uncertainty error metric \cite{miller2019evaluating} to assess the model's ability to distinguish correct and incorrect detections based on the uncertainty estimate threshold ($\delta$).
\begin{equation}
\label{eq:ue}
    TPRj = \frac{|U(\mathbbm{D}_c) > \delta|}{|\mathbbm{D}_c|}, \quad FPRt = \frac{|U(\mathbbm{D}_i) \leq \delta|}{|\mathbbm{D}_i|}, \quad UE = \frac{TPRj + FPRt}{2}
\end{equation}
\begin{equation}
\begin{aligned}
        \mathcal{U}_{pred} = 1 - \hat{C}
\end{aligned}
\label{eq:upred}
\end{equation}
where \textit{TPRj} (True Positive Rejection) is a proportion of correct detections $\mathbbm{D}_c$ which are incorrectly rejected ($U$ is the uncertainty measure of detections $\mathbbm{D}$), and \textit{FPRt} (False Positive Retention) is a proportion of incorrect detections $\mathbbm{D}_i$ which are incorrectly accepted. The optimal value of $\mathbf{\delta}$ is chosen, giving maximum true positives and minimum false positives out of all combined detections. A perfect uncertainty error score of 0\% indicates that all incorrect detections ($\mathbbm{D}_i$) are rejected, and all correct detections ($\mathbbm{D}c$) are accepted. Both the true positive rate and the false positive rate are equally weighted. Since the BEA incorporates Tandem Loss functions that calibrate the model's confidence score, the predicted object uncertainty $\mathcal{U}_{pred}$ can be inferred directly as the complement of the confidence score.
\textbf{Detection Accuracy (Average Precision)}
We evaluate the accuracy of our object detection models using two variants of the Average Precision (AP) metric: mAP and AP50, which measure the mean AP over different IOU thresholds ([50:5:95]) and the AP at a 50\% intersection over the union (IOU) threshold, respectively. We calculate $\mathbf{AP50_{\mathcal{U}_{raw}}}$ and $\mathcal{U}_{pred}$-based $\mathbf{AP50_{\mathcal{U}_{pred}}}$ scores to demonstrate the effectiveness of the BEA approach. We compute the raw AP50 by applying post-NMS predictions using a confidence threshold of 0.05. For the $\mathcal{U}_{pred}$-based AP50 ($\mathbf{AP50_{\mathcal{U}_{pred}}}$), we exclude samples with $\mathcal{U}_{pred}$ $>\delta$. A well-calibrated model with precise uncertainty estimates will likely have AP50 scores closer to the raw AP50.

\textbf{AP50-based Retention Curves:} 
\label{sec:retention_curve}
Inspired by the Shifts benchmark \cite{malinin2021shifts}, we adopt retention curves, commonly used for regression tasks, to assess the object detection model's robustness and quality of the calibration. 
Reliability diagrams ~\cite{degroot1983comparison} are unsuitable for object detection models as they are primarily used to measure the calibration of confidence scores in classification tasks. Retention curves are better suited for the complex AP metric involving regression and classification. Specifically, we use AP50-based retention curves, which involve sorting predictions by their descending uncertainty level and calculating AP50 scores at various fractions of retained predictions as  illustrated in Fig. ~\ref{fig:ap_reten}. The AUC of these curves measures the calibration quality of confidence scores. A poorly calibrated model will need a larger fraction (x-axis) to attain the $\mathbf{AP50_{\mathcal{U}_{raw}}}$. This metric should only compare the model's calibration using the same validation dataset to avoid bias.

\textbf{Retention Curves vs. Uncertainty Error}: To comprehensively evaluate object detection models, consider both AP50-based retention curves and uncertainty error metrics. Note that models with fewer detections may have lower UE metric values for \textit{TPRj} or \textit{FPRt} (eq. ~\ref{eq:ue}), indirectly reducing the UE metric. Precision-recall curves may not penalize false positives with low confidence scores, affecting AP50-based retention curves \cite{qutub2022hardware}.

\textbf{Out of distribution (OOD) detection:} 
The uncertainty measure $\mathcal{U}_{ood}$ is calculated by taking the mean ($\mathcal{\mu}$) of the entropy value for each grid cell and anchors from the prediction of both detectors, as shown in eq.~\ref{eq:OOD}. This metric determines whether input data is in-distribution or out-of-distribution by quantifying whether out-of-distribution input is correctly detected, i.e., assigned a high uncertainty. In contrast, in-distribution data should be assigned a low uncertainty value. To evaluate the effectiveness of the uncertainty measure $\mathcal{U}_{ood}$ in identifying out-of-distribution data, we compute the AUC-ROC \cite{bradley1997use} using the $\mathcal{U}_{ood}$ values across a complete dataset that combines in-distribution and out-of-distribution data in a 1:2 ratio. Using the BEA property with $\mathcal{L_{\mathbf{tandem}}}$, we combine the mean squared error of tandem bounding box prediction (\^x, \^y, \^w, and \^h), confidence score, and entropy from $\mathcal{\alpha}$ and $\mathcal{\beta}$ detectors as an uncertainty measure $\mathcal{U}_{ood}$, as shown in eq.\ref{eq:OOD} and eq.\ref{eq:OOD_final}. The $\mathcal(s, b)$ in eq.\ref{eq:OOD} and eq.\ref{eq:OOD_final} refers to all anchors which are accessed individually.
\begin{equation}
\centering
\label{eq:OOD}
\begin{aligned}
\scalebox{0.9}{$
\mathfrak{B}( z, i, j) =  \left(\hat{z}^{\mathcal{\alpha}}_{ij} - \hat{z}^{\mathcal{\beta}}_{ij}\right)^2,  \quad
\mathcal{B}(s, b) =  \sqrt{\sum_{z \in (\hat{x}, \hat{y}, \hat{w}, \hat{h}, \hat{C})}\mathfrak{B}( z, s, b)}, \quad
H(X) = -\sum_{i=1}^{n}p(x_{i})\ln(p(x_{i}))$,} \\
\scalebox{0.9}{$\mathfrak{h}( z, i, j) =  \left(H(\hat{z}^{\mathcal{\alpha}}_{ij}) - H(\hat{z}^{\mathcal{\beta}}_{ij})\right)^2, \quad
\mathcal{H}(s, b) =  \sqrt{\sum_{z \in \hat{c}}\mathfrak{h}( z, s, b)}.
$}
\end{aligned}
\end{equation}
\begin{equation}
\label{eq:OOD_final}
\begin{aligned}
\mathcal{U}_{ood} &= \mathcal{\mu} \Bigl( \mathcal{B}(s, b)*\mathcal{H}(s, b)\Bigl)
\end{aligned}
\end{equation}
For non-BEA-YOLOv3 and non-Gaussian-based YOLOv3 models, we utilize their confidence scores as their $\mathcal{U}_{ood}$. However, for Gaussian-based YOLOv3 models, we use its $Uncertainty_{aver}$ as $\mathcal{U}_{ood}$. To assess the OOD detection performance of the models, we use two near-OOD datasets $\mathcal{U}_{near-ood}$ (CityPersons \cite{zhang2017citypersons}, and BDD100K \cite{yu2020bdd100k}) and one far-OOD dataset $\mathcal{U}_{far-ood}$ (COCO \cite{lin2014microsoft}) and compute the AUC-ROC values. Higher AUC-ROC values indicate better OOD detection performance.
\subsection{Results and Discussion}

\begin{table*}[tbp]
\centering
\captionsetup{justification=centering}
\resizebox{0.95\textwidth}{!}{%
\begin{tabular}{|c|c|cc|c|c|ccc|}

\hline
\multirow{2}{*}{\textbf{\begin{tabular}[c]{@{}c@{}}Models \\ (input size 416$\times$416)\end{tabular}}} & \multirow{2}{*}{$\mathbf{mAP_{raw}}$ (\%) $\mathbf{\uparrow}$} & \multicolumn{2}{c|}{$\mathbf{AP50\; \mathbf{\uparrow}}$} & \multirow{2}{*}{\textbf{UE (\%) $\mathbf{\downarrow}$}} & \multirow{2}{*}{\textbf{\begin{tabular}[c]{@{}c@{}}AP50-based Retention curve \\ AUC (\%) $\mathbf{\uparrow}$\end{tabular}}} & \multicolumn{3}{c|}{\textbf{\begin{tabular}[c]{@{}c@{}}Out-of-distribution detection (OOD) \\ AUC-ROC (\%)$\mathbf{\uparrow}$ \end{tabular}}} \\ 
\cline{3-4} \cline{7-9} & & \multicolumn{1}{c|}{$\mathbf{AP50_{raw}}$} & $\mathbf{AP50_{\mathcal{U}_{pred}}}$ &  &  & \multicolumn{1}{c|}{\textbf{\begin{tabular}[c]{@{}c@{}}CityPersons \\$\mathcal{U}_{near-ood}$ \end{tabular}}} & \multicolumn{1}{c|}{\textbf{\begin{tabular}[c]{@{}c@{}}BDD100K\\ $\mathcal{U}_{near-ood}$\end{tabular}}} & \textbf{\begin{tabular}[c]{@{}c@{}}COCO\\$\mathcal{U}_{far-ood}$\end{tabular}} \\ \hline \hline
Base-YOLOv3 & 51.72 & \multicolumn{1}{c|}{87.4} & 78.2 & 11.96 & 53.1 & \multicolumn{1}{c|}{35*} & \multicolumn{1}{c|}{40.16*} & 20.21 \\ \hline
\begin{tabular}[c]{@{}c@{}}YOLOv3\\ 3 Ensemble\end{tabular} & 54.58 & \multicolumn{1}{c|}{89} & 82.94 & 9.23 & 58.7 & \multicolumn{1}{c|}{28.79*} & \multicolumn{1}{c|}{32.44*} & 20.5* \\ \hline
\begin{tabular}[c]{@{}c@{}}YOLOv3\\  5 Ensemble\end{tabular} & \underline{\textbf{55.1}} & \multicolumn{1}{c|}{89.27} & 82.97 & 9.03 & 59.3 & \multicolumn{1}{c|}{28.6*} & \multicolumn{1}{c|}{12.19*} &  10.21*\\ \hline
\textbf{\begin{tabular}[c]{@{}c@{}}BEA-YOLOv3\end{tabular}} & \textbf{54.83 $\pm$ 0.28} & \multicolumn{1}{c|}{\underline{\textbf{89.3 $\pm$ 0.28}}} & \underline{\textbf{85.79 $\pm$ 0.13}} & \underline{\textbf{4.55 $\pm$ 0.02}} & \underline{\textbf{73.9 $\pm$ 1.1}} & \multicolumn{1}{c|}{\underline{\textbf{98.75 $\pm$ 0.3}}} & \multicolumn{1}{c|}{\underline{\textbf{86.71 $\pm$ 1.7}}} & \underline{\textbf{97.33 $\pm$ 0.9}} \\ \hline \hline
Gaussian YOLOv3 & 47.65 & \multicolumn{1}{c|}{88.17} & 83.98 & 4.96 & 81.8 & \multicolumn{1}{c|}{78.98**} & \multicolumn{1}{c|}{67.49**} & 91.33** \\ \hline
\begin{tabular}[c]{@{}c@{}}Gaussian YOLOv3\\ +\\  3 Ensemble\end{tabular} & 50.43 & \multicolumn{1}{c|}{89.48} & 85.66 & 4.61 & 84.4 & \multicolumn{1}{c|}{81.72**} & \multicolumn{1}{c|}{71.08**} & 89** \\ \hline
\begin{tabular}[c]{@{}c@{}}Gaussian YOLOv3\\ +\\ 5 Ensemble\end{tabular} & 52.29 & \multicolumn{1}{c|}{89.92} & 85.79 & 4.55 & 84.7 & \multicolumn{1}{c|}{82.31**} & \multicolumn{1}{c|}{71.56**} & 84.8** \\ \hline
\textbf{\begin{tabular}[c]{@{}c@{}}BEA-Gaussian-YOLOv3\end{tabular}} & \underline{\textbf{54.28 $\pm$ 0.14}} & \multicolumn{1}{c|}{\underline{\textbf{90.64 $\pm$ 0.34}}} & \underline{\textbf{86.5 $\pm$ 0.1}} & \underline{\textbf{4.05 $\pm$ 0.01}} & \underline{\textbf{86.2 $\pm$ 0.4}} & \multicolumn{1}{c|}{\underline{\textbf{79.21 $\pm$ 1.7}}} & \multicolumn{1}{c|}{\underline{\textbf{85.89 $\pm$ 3}}} & \underline{\textbf{98.4 $\pm$ 1.1}} \\ \hline \hline

\begin{tabular}[c]{@{}c@{}}Base-SSD\end{tabular} & 51.24 & \multicolumn{1}{c|}{88.69} & 80.15 & 11.28 & 73.5 & \multicolumn{1}{c|}{42.95*} & \multicolumn{1}{c|}{45.41*} & 26.37* \\ \hline

\begin{tabular}[c]{@{}c@{}}SSD\\ +\\ 5 Ensemble\end{tabular} & 52.8 & \multicolumn{1}{c|}{89.14} & 82.71 & 10.06 & 78.6 & \multicolumn{1}{c|}{38*} & \multicolumn{1}{c|}{41.28*} & 37.8* \\ \hline

\textbf{\begin{tabular}[c]{@{}c@{}}BEA-SSD\end{tabular}} & \underline{\textbf{53.18 $\pm$ 0.08}} & \multicolumn{1}{c|}{\underline{\textbf{90.38 $\pm$ 0.17}}} & \underline{\textbf{86.83 $\pm$ 0.4}} & \underline{\textbf{7.7 $\pm$ 0.08}} & \underline{\textbf{82.5 $\pm$ 0.4}} & \multicolumn{1}{c|}{\underline{\textbf{61.3 $\pm$ 2.1}}} & \multicolumn{1}{c|}{\underline{\textbf{61.38 $\pm$ 3.4}}} & \underline{\textbf{88.49 $\pm$ 0.87}} \\ \hline
\end{tabular}%
}
\caption{Evaluation of the BEA architecture on YOLOv3 and SSD using various metrics, including accuracy, uncertainty estimation, calibration, robustness, and retention curves. Trained on all seven usable classes of KITTI with a constant set of hyperparameters, the test set comprised 15\% of KITTI data. The effectiveness of OOD detection was demonstrated using one far-ood and two near-ood datasets. * represents that the confidence score is used for detecting OOD data, and ** represents Gaussian uncertainty is used for the same purpose.}
\label{tab:bea_results_table}
\end{table*}

BEA improves both YOLOv3 and SSD model's calibration and prediction accuracy with low uncertainty error (UE), enabling it to detect out-of-distribution data. The model's accuracy is evaluated using the average precision metric, which requires a minimum overlap of 50\%, but a minimum confidence score of 0.05 can lead to higher false positives. An accurate uncertainty estimation model can identify and discard samples with high uncertainty scores, resulting in lower UE. This section discusses Table ~\ref{tab:bea_results_table}, particularly the YOLOv3 results in detail. Table ~\ref{tab:bea_results_table} shows that YOLOV3 with BEA has the lowest UE compared to Base-YOLOv3, while Gaussian-YOLOv3 has a better UE (around 4.9\%). Incorporating BEA into Gaussian-YOLOv3 further reduces UE to approximately 4\%, demonstrating BEA's significant improvement in uncertainty estimation, regardless of the detection model's conventional loss functions.

BEA improves YOLOv3 accuracy with $mAP_{raw}$ and $AP50_{raw}$ increasing by 6\% and 3.7\% respectively over base-YOLOv3. BEA-Gaussian-YOLOv3 shows 14\% and 2.8\% improvement of $mAP_{raw}$ and $AP50_{raw}$ over Gaussian-YOLOv3. The Gaussian-modeled loss function used in Gaussian YOLOv3 enables it to excel at learning from more evidence and perform better on dominant classes. However, Gaussian YOLOv3 may not perform well for less prominent classes, leading to a lower-than-expected $mAP_{raw}$ when evaluated with all seven classes of the KITTI dataset. Discarding predictions with high uncertainty scores using $\mathcal{U}{pred}$ has little impact on BEA's $AP50{\mathcal{U}_{pred}}$, indicating more accurate uncertainty estimates. BEA achieves approximately 9.6\% higher AP50 than base-YOLOv3 and 4.2\% higher AP50 than Gaussian-YOLOv3 using these uncertainties. The uncertainty measure $AP50_{\mathcal{U}_{pred}}$ considers the overall uncertainty of a prediction, as the Tandem Loss functions work collectively to reduce variance using all factors of object prediction. Our uncertainty measure doesn't separate uncertainty into spatial or localization uncertainty. Our Gaussian-YOLOv3 implementation has a lower uncertainty error than Gasperini et al. \cite{gasperini2021certainnet}.
\begin{figure}
\centering
\begin{tabular}{cccc}
\bmvaHangBox{\fbox{\includegraphics[width=2.6cm, height=2cm]{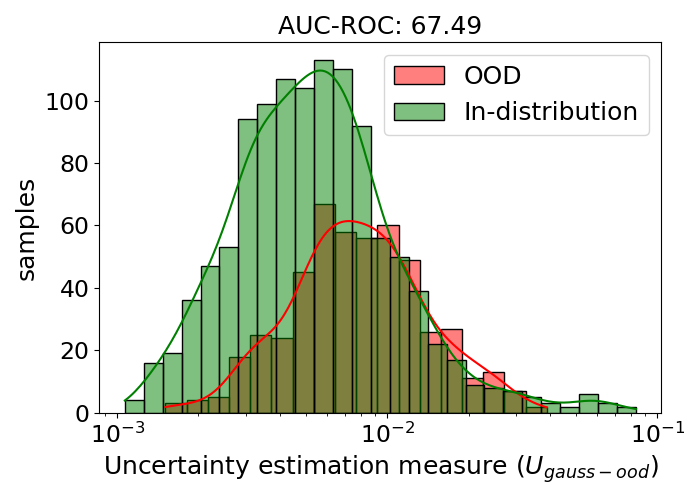}}}&
\bmvaHangBox{\fbox{\includegraphics[width=2.6cm, height=2cm]{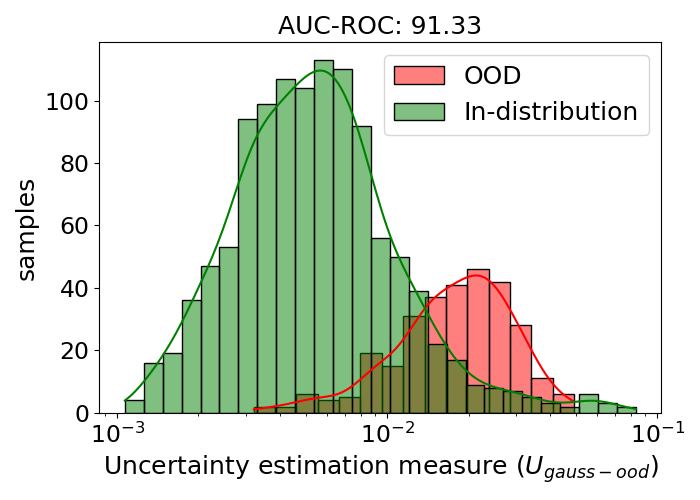}}}&
\bmvaHangBox{\fbox{\includegraphics[width=2.6cm, height=2cm]{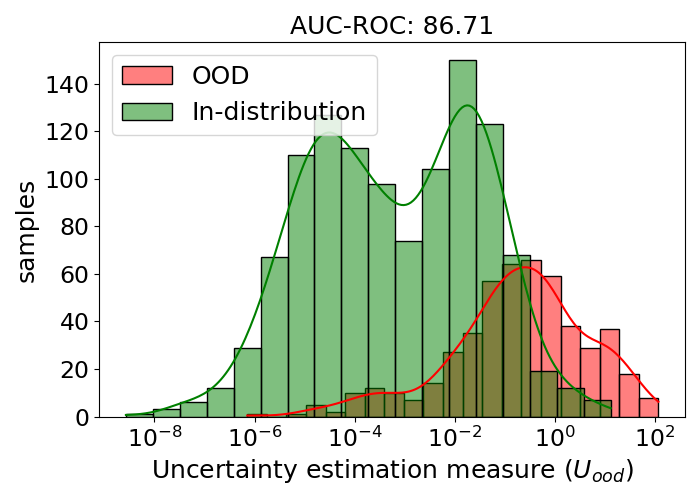}}}&
\bmvaHangBox{\fbox{\includegraphics[width=2.6cm, height=2cm]{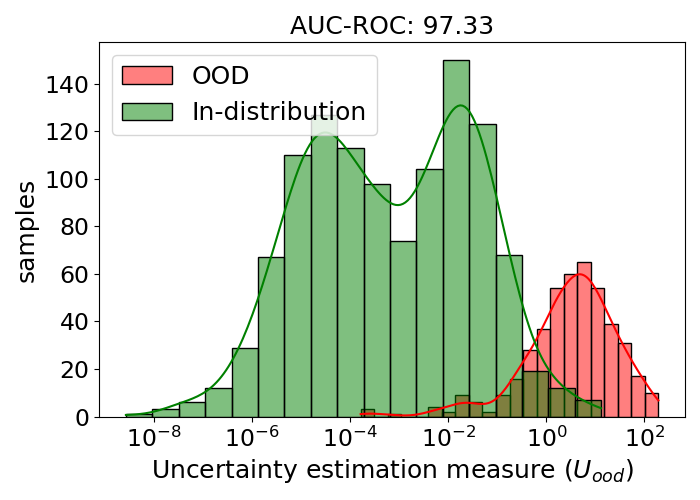}}}\\
(a)&(b)&(c)&(d)
\end{tabular}
\caption{Out-of-distribution image detection with KITTI-trained YOLOv3 models on COCO and BDD1000K datasets: (a) Gauss-YOLOv3 - BDD100K; (b) Gauss-YOLOv3 - COCO; (c) BEA-YOLOv3 - BDD100K; (d) BEA-YOLOv3 - COCO}
\label{fig:ood_gauss_bdd}
\label{fig:ood_gauss_coco}
\label{fig:ood_BEA_bdd}
\label{fig:ood_fig}
\end{figure}

\begin{wrapfigure}{r}{0.5\textwidth}
    \centering
    \bmvaHangBox{\fbox{\includegraphics[width=0.5\textwidth, scale=0.5]{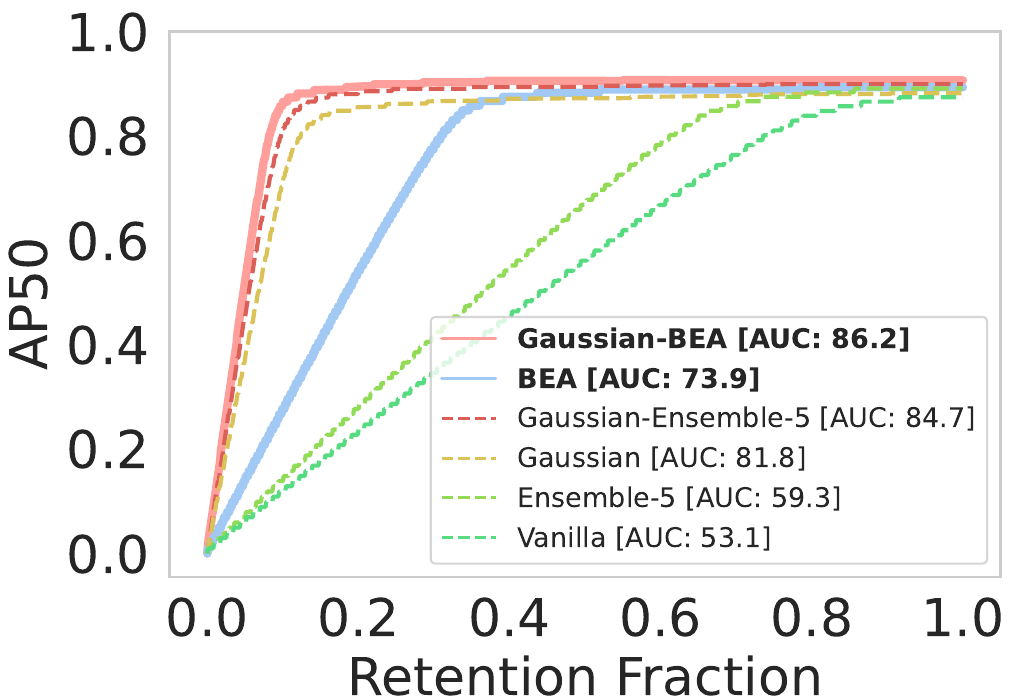}}}
    \caption{AP50-based retention curves}
\label{fig:ap_reten}
\end{wrapfigure}

Fig. ~\ref{fig:ap_reten} displays the AP50 Retention curves for Base-YOLOv3, 5-member ensemble, Gaussian-YOLOv3, and BEA models. The findings reveal that BEA-YOLOv3 performs better than Base-YOLOv3, with a 40\% increase in the area under the AP50-based retention curve. Additionally, BEA Gaussian-YOLOv3 shows a 5.4\% enhancement over Gaussian-YOLOv3, demonstrating improved calibration of confidence scores. BEA's effectiveness is shown to depend on the base loss functions and their ability to train $\alpha$ and $\beta$ detectors to learn representations independently, as observed in the significant improvement in the UE results when BEA is incorporated into the Gaussian version.
Additionally, BEA's aggregation approach, as discussed in Section ~\ref{sec:bea_on_yolov3}, contributes to its superior performance. The \textbf{Tandem Loss} functions improve confident predictions at each tandem pair during training. Disagreements in negative predictions between bounding boxes, confidence scores, and objectness scores can identify out-of-distribution (OOD) data when detectors disagree with their corresponding tandem detector $\beta$.

BEA's OOD detection ability is evaluated by combining in-distribution with out-of-distribution datasets at a 2:1 ratio of 1000 and 500 validation images. 
The BEA model performs better than all other models in OOD detection ability, as shown in Table ~\ref{tab:bea_results_table} and Fig. ~\ref{fig:ood_fig}. AUC-ROC is calculated by averaging the OOD uncertainty scores $\mathcal{U}_{ood}$ for all predictions made on each image. The model is evaluated with two near-ood datasets (Citypersons \cite{zhang2017citypersons} and BDD100K \cite{yu2020bdd100k}) and the in-distribution dataset. 

The BEA model exhibits superior performance compared to all other models, with a significant margin of improvement. BEA has a lower FPS than Base-YOLOv3 due to sequential execution, but parallelizing tandem layers could improve this. Despite having fewer parameters than ensembles, BEA surpasses them in accuracy, calibration, and uncertainty estimates.


\section{Conclusion and Future Work} In conclusion, this study proposes a novel Budding-Ensemble Architecture (BEA) for Single-Stage Anchor-based models, which includes a new loss function called tandem loss functions. The proposed architecture enhances detection accuracy and well-calibrated confidence scores leading to higher-quality uncertainty estimates. Evaluation results using various metrics demonstrate the superiority of BEA over base models (YOLOv3 and SSD) and ensembles regarding calibration and uncertainty estimation. The architecture's unique structural advantage provides an opportunity to enhance model reliability in different settings. Future work can optimize and integrate BEA into vision-based models, potentially replacing ensembles for improved accuracy and efficiency. Additionally, future work will study the architecture split point where detectors are duplicated to understand its impact on score calibration and OOD detection. Additionally, BEA's effectiveness can be explored in more complex scenarios, such as object detection in crowded or cluttered scenes. Overall, BEA is a promising framework for improving object detection model reliability and optimizing its integration can lead to significant improvements in model performance on various tasks.

\textbf{Acknowledgments}: This work was partially funded by the Federal Ministry for Economic Affairs and Climate Action of Germany, as part of the research project SafeWahr (Grant Number: 19A21026C) 


\bibliography{bibliography}
\newpage 
\section{Supplementary Material: BEA}

\subsection{Evaluation of Tandem Loss on Confidence Prediction in BEA using YOLOv3:}

\begin{figure}
\centering
\begin{tabular}{cc}
\bmvaHangBox{\fbox{\includegraphics[width=5.9cm, height=4cm]{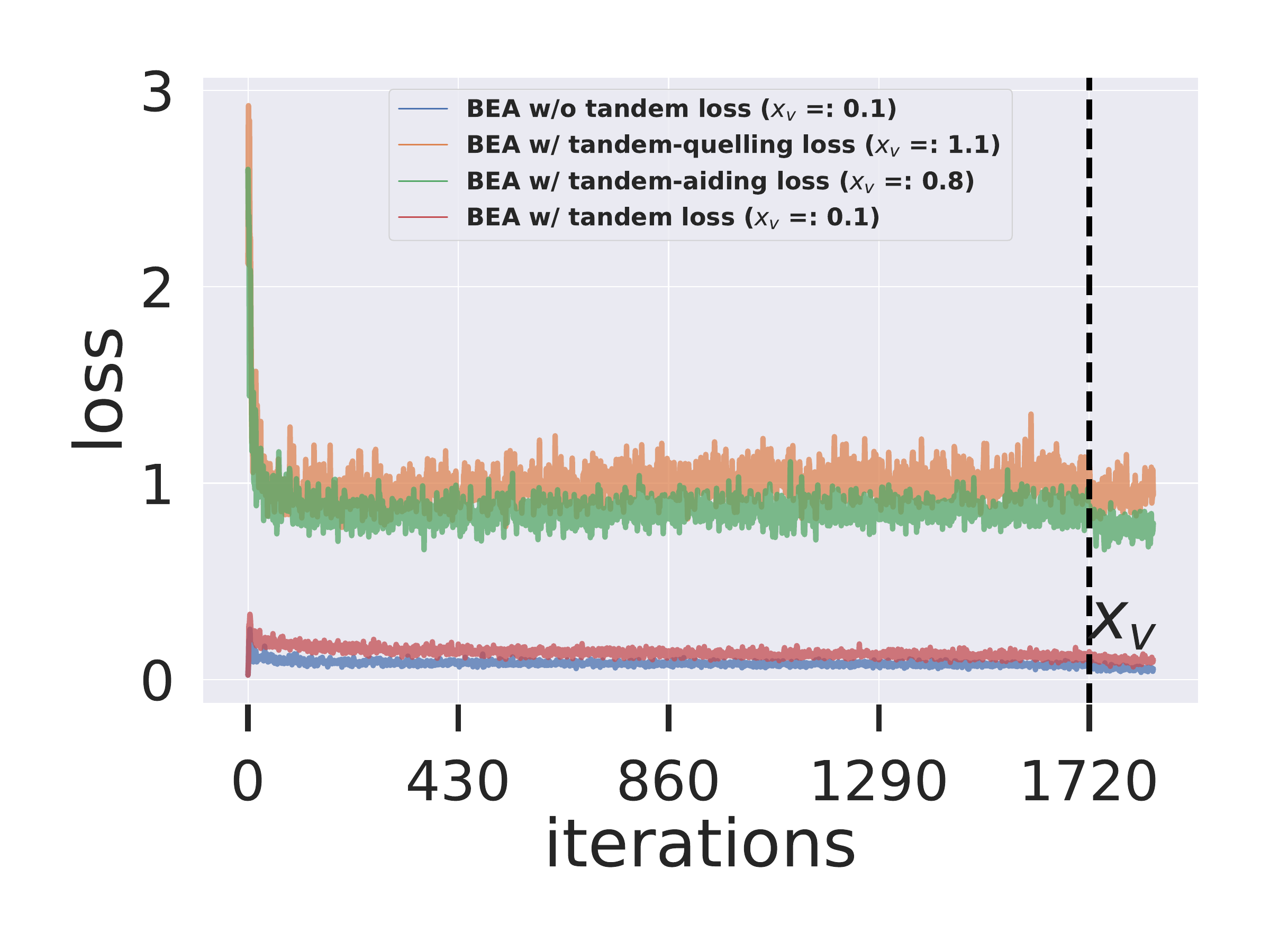}}}&
\bmvaHangBox{\fbox{\includegraphics[width=5.9cm, height=4cm]{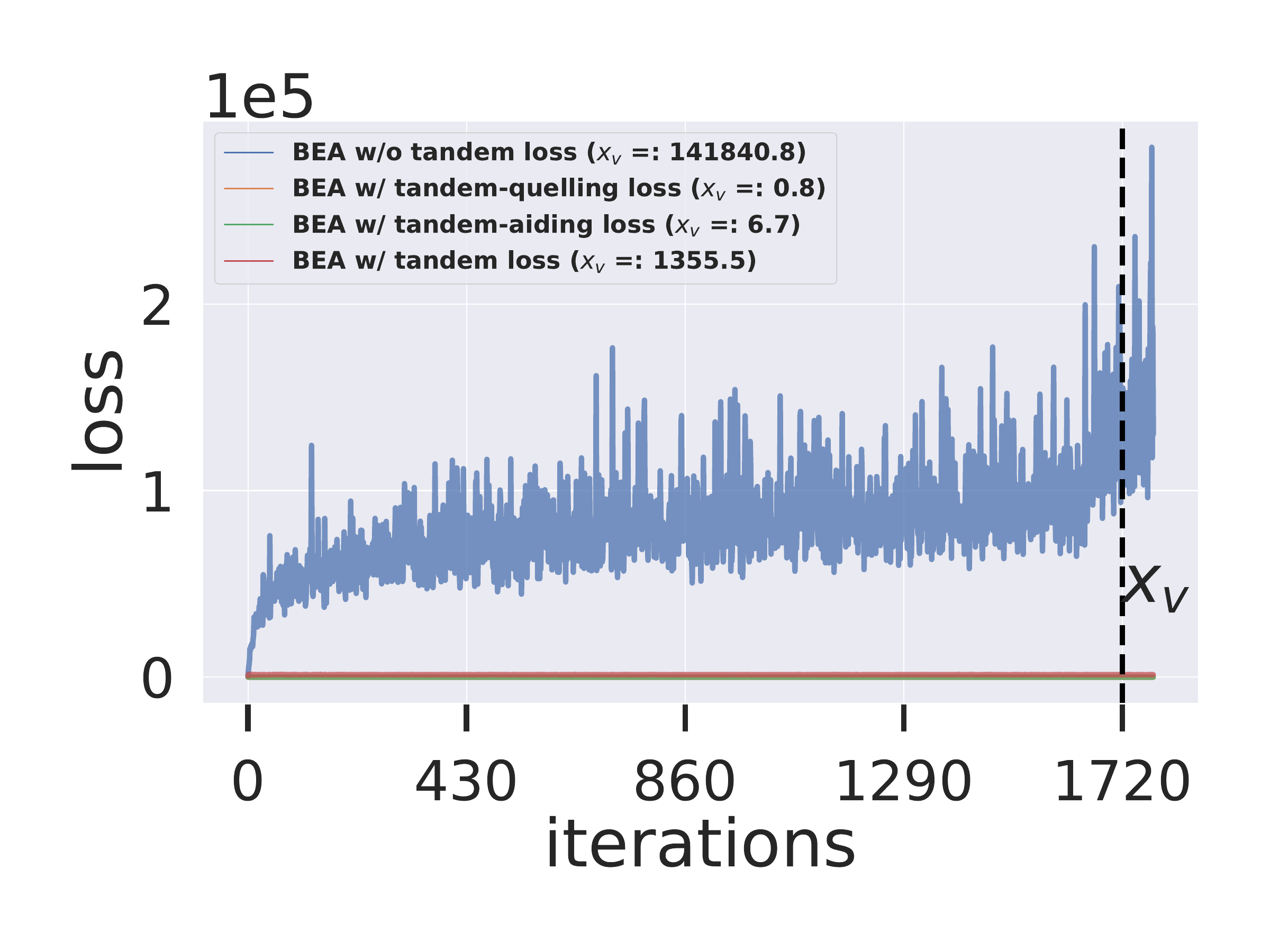}}}\\
(a)&(b)
\end{tabular}
\caption{Monitoring objectness-based losses - Impact of Tandem Loss ($\mathcal{L_{\mathbf{tandem}}}$) across various BEA-YOLOv3 configurations) : (a) Monitoring of $\mathcal{L_{\mathbf{ta}}}$  on predicted confidence scores for positive predictions from the Tandem Detectors; (b) Monitoring of $\mathcal{L_{\mathbf{tq}}}$  on predicted confidence scores for negative predictions from the Tandem Detectors}
\label{fig:obj-loss-pos}
\label{fig:obj-loss-neg}
\label{fig:obj-loss}
\end{figure}

Figure ~\ref{fig:obj-loss} illustrates the impact of Tandem Loss-$\mathcal{L_{\mathbf{tandem}}}$ on confidence prediction loss, which is one of the regression losses of YOLOv3. The experiment was conducted on Budding-Ensemble Architecture (BEA) to evaluate the usability of Tandem Loss. The loss of predicted confidence score reflects the model's proficiency in detecting an object by a particular anchor.
Figure ~\ref{fig:obj-loss-pos} displays the monitoring of $\mathcal{L_{\mathbf{ta}}}$, while Figure ~\ref{fig:obj-loss-neg} displays the monitoring of $\mathcal{L_{\mathbf{tq}}}$ losses. The application and monitoring of these losses depend on the configuration indicated in the legends of the corresponding figures. Figure~\ref{fig:obj-loss-pos} demonstrates that the BEA with Tandem Loss configuration shows superior performance in reducing the variance of respective positive predictions between $\alpha$ and $\beta$ detectors by decreasing the $\mathcal{L_{\mathbf{ta}}}$ compared to the separate factors of $\mathcal{L_{\mathbf{tandem}}}$ loss, including the absence of $\mathcal{L_{\mathbf{tandem}}}$ loss. This is also shown by the data points along a vertical line denoted by $X_{V}$ in the legends. 
Similarly, the Figure in ~\ref{fig:obj-loss-neg} illustrates that the BEA without the Tandem Loss results in the decreased variance of negative predictions between the $\alpha$ and $\beta$ detectors, which is not desirable as we prefer to have high variance between the corresponding negative predictions. In Figure ~\ref{fig:obj-loss-neg}, it is worth noting that the BEA configuration with individual factors of $\mathcal{L_{\mathbf{tandem}}}$ loss produces a preferred result in increasing the variance between negative predictions. However, this also hurts $\mathcal{L_{\mathbf{ta}}}$ loss as shown in Figure ~\ref{fig:obj-loss-pos}. This indicates that using both $\mathcal{L_{\mathbf{ta}}}$ and $\mathcal{L_{\mathbf{tq}}}$ together leads to better performance in improving the confidence scores of True Positives and reducing the scores of False Positives. This results in better-calibrated prediction outcomes.

\subsection{Extended Ablation Study of Table ~\ref{tab:bea_results_table} with OOD results on YOLOv3}

\begin{table*}[hbt!]
\resizebox{\linewidth}{!}{
\centering
\begin{tabular}{|cc|c|c|c|c|c|ccc|}
\hline
 \multicolumn{2}{|c|}{\begin{tabular}[c]{@{}c@{}}$\mathcal{L_{\mbox{tandem}}}$\end{tabular}} & \multirow{2}{*}{$mAP_{raw}$ $\mathbf{\uparrow}$} & \multirow{2}{*}{$AP50_{raw}$ $\mathbf{\uparrow}$} & \multirow{2}{*}{$AP50_{\mathcal{U}_{pred}}$ $\mathbf{\uparrow}$} & \multirow{2}{*}{$UE$ (\%) $\mathbf{\downarrow}$} & \multirow{2}{*}{\begin{tabular}[c]{@{}c@{}}Retention Curve\\ $AUC$ (\%) $\mathbf{\uparrow}$\end{tabular}} & \multicolumn{3}{|c|}{\begin{tabular}[c]{@{}c@{}}Out-of-distribution detection (OOD) \\ AUC-ROC (\%)$\mathbf{\uparrow}$ \end{tabular}} \\
 \cline{1-2} \cline{8-10} \multicolumn{1}{|c|}{$\mathcal{L_{\mbox{ta}}}$} & $\mathcal{L_{\mbox{tq}}}$ &  & & & & & \multicolumn{1}{c|}{\begin{tabular}[|c|]{@{}c@{}}CityPersons \\$\mathcal{U}_{near-ood}$ \end{tabular}} & \multicolumn{1}{c|}{\begin{tabular}[c]{@{}c@{}}BDD100K\\ $\mathcal{U}_{near-ood}$\end{tabular}} & \begin{tabular}[c]{@{}c@{}}COCO\\$\mathcal{U}_{far-ood}$\end{tabular} \\ \hline
\multicolumn{1}{|c|}{\xmark} & \xmark &  52.36 & 87.96   &  79.11 & 9.65 & 56.9& \multicolumn{1}{c|}{81.05} & \multicolumn{1}{c|}{77.21} & 94.81  \\ \hline
\multicolumn{1}{|c|}{\cmark} & \xmark &  54.07 & 88.56   & 79.44 & 11.05 &  54.2& \multicolumn{1}{c|}{77.01} & \multicolumn{1}{c|}{75.97} & 93.63\\ \hline
\multicolumn{1}{|c|}{\xmark} & \cmark &  54.82 &  88.31  & 82.15 &  9.03 &  57.1& \multicolumn{1}{c|}{72.8} & \multicolumn{1}{c|}{73.79} & 91.61 \\ \hline
\multicolumn{1}{|c|}{\cmark} & \cmark &  \textbf{54.83}  &  \textbf{89.2} & \textbf{85.79} & \textbf{4.55}  &  \textbf{73.9} & \multicolumn{1}{c|}{\textbf{98.75}} & \multicolumn{1}{c|}{\textbf{86.71}} & \textbf{97.33} \\ \hline
\end{tabular}%
}
\caption{Extended Ablation study on BEA-YOLOv3 with on YOLOv3: Effects on the accuracy, uncertainty error, calibration and out-of-distribution shown by $\mathcal{L_{\mbox{tandem}}}$.}
\label{tab:supp_ablation_bea}
\end{table*}

Table ~\ref{tab:supp_ablation_bea} provides an extended ablation study to understand the dependency of $\mathcal{L_{\mathbf{ta}}}$ and $\mathcal{L_{\mathbf{tq}}}$ loss functions individually. This comprehensive study presents out-of-distribution (OOD) results for various configurations of Tandem Loss ($\mathcal{L_{\mathbf{tandem}}}$). The table highlights the importance of adding $\mathcal{L_{\mathbf{ta}}}$ and $\mathcal{L_{\mathbf{tq}}}$ Tandem Loss together leading to improved prediction accuracy and calibration of confidence score, higher uncertainty estimates, and better OOD detection performance.

\begin{table}
\centering
\begin{tabular}{|c|c|c|}
\hline
Model & Parameters (MM) \\ \hline
Base-YOLOv3 & 61.5 \\ \hline
M-Ensemble YOLOv3 & M$\times$61.5 \\ \hline
BEA-YOLOv3 &  82.53 \\ \hline \hline
Base-SSD & 25.5 \\ \hline
M-Ensemble SSD & M$\times$25.5 \\ \hline
BEA-SSD & 30.63 \\ \hline
BEA-SSD Arch2 & 27.5 \\ \hline
\end{tabular}%
\caption{Computational-Overhead Analysis of BEA}
\label{tab:overhead}
\end{table}

\begin{figure}
\centering
\bmvaHangBox{\fbox{\includegraphics[width=0.9\linewidth]{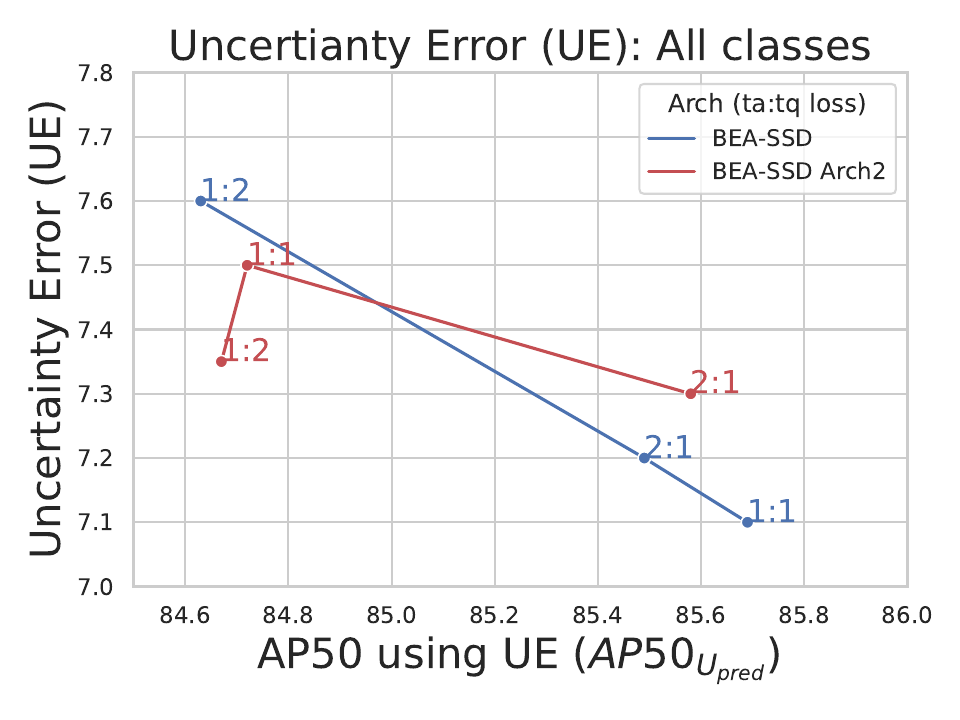}}}
\caption{Analysing different versions of BEA-SSD to optimise the computational overhead}
\label{fig:SSD-overhead}
\end{figure}

\subsection{SSD architecture with BEA:}

\begin{figure}
\centering
\begin{tabular}{c}
\bmvaHangBox{\fbox{\includegraphics[width=0.9\linewidth]{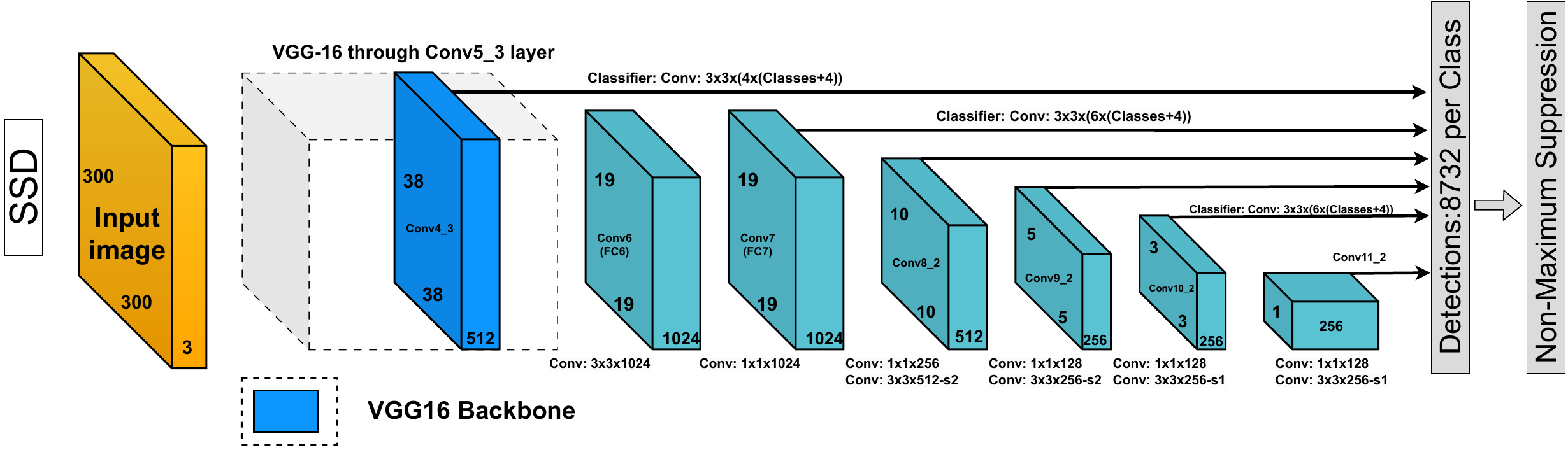}}}\\
(a)\\
\bmvaHangBox{\fbox{\includegraphics[width=0.9\linewidth]{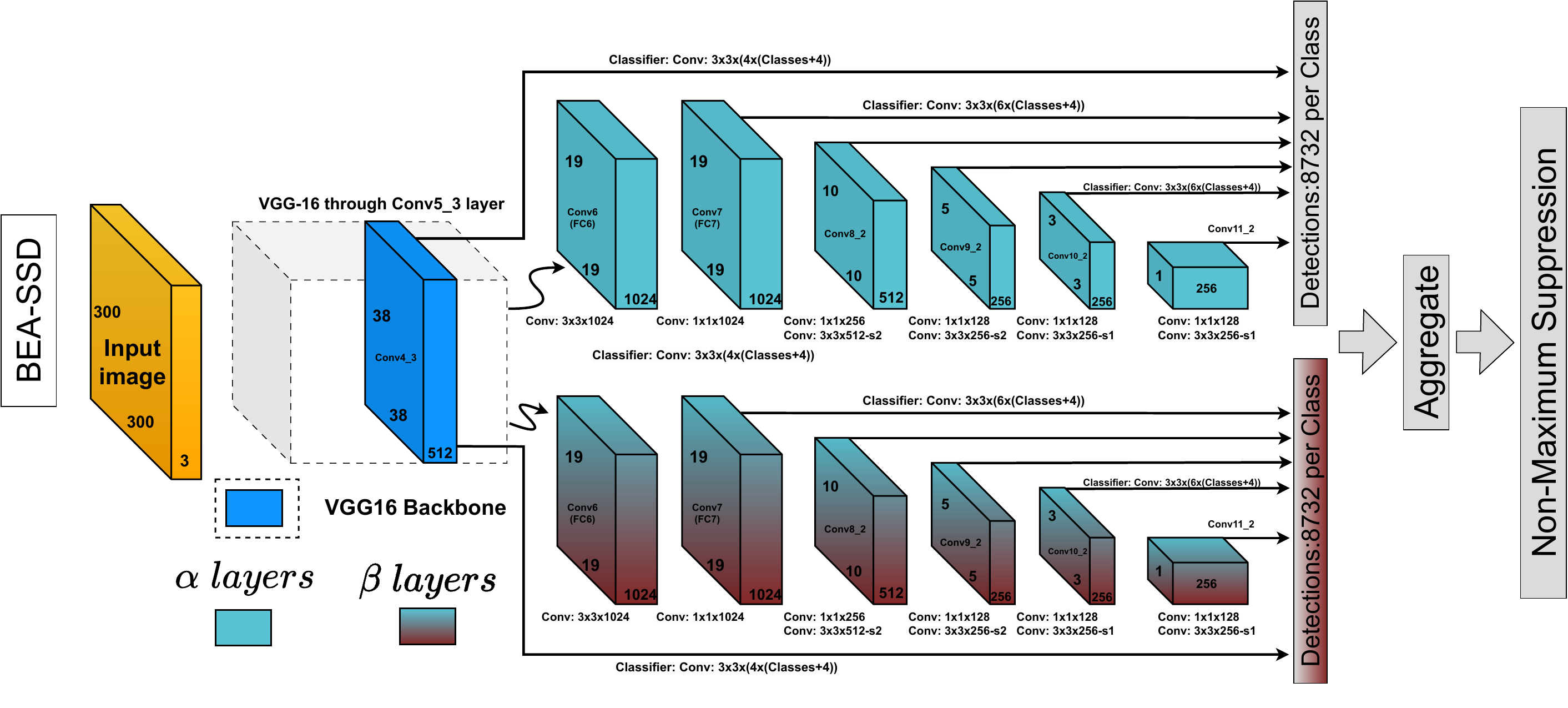}}}\\
(b)\\
\bmvaHangBox{\fbox{\includegraphics[width=0.9\linewidth]{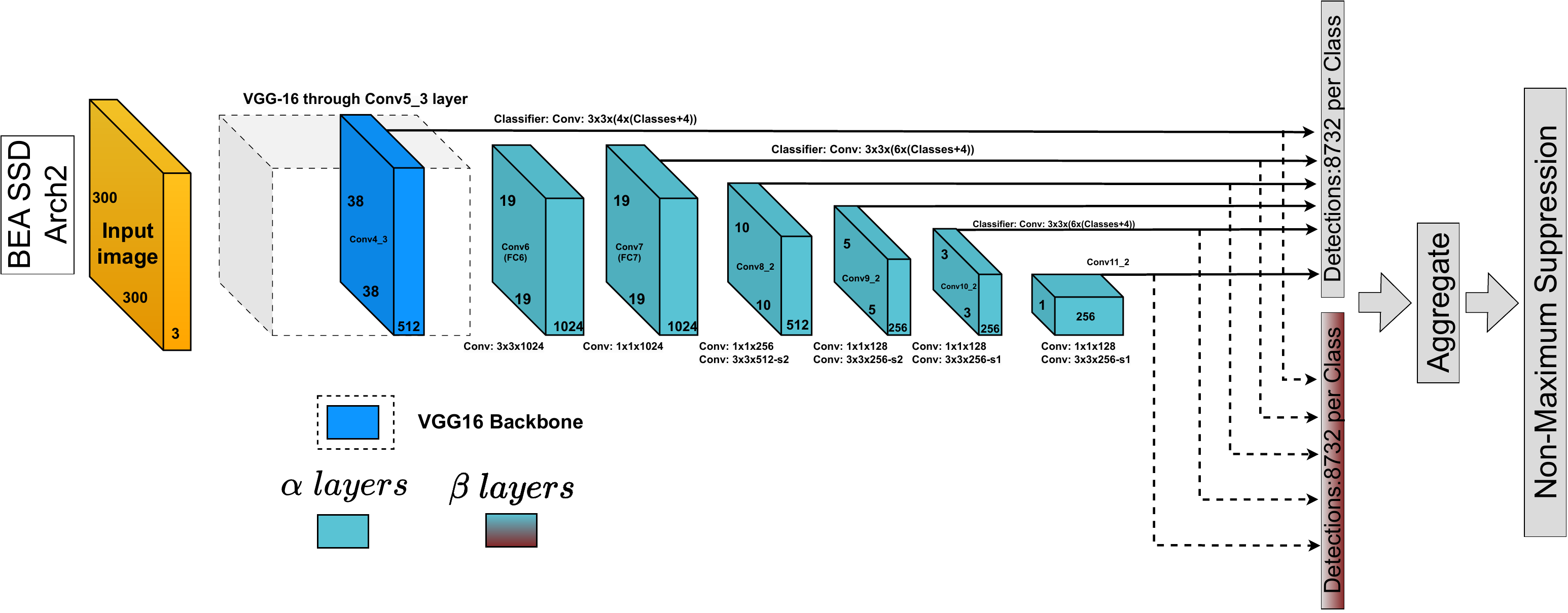}}}\\
(c)
\end{tabular}
\caption{Different versions of SSD : (a) Base-SSD (Vanilla version); (b) BEA-SSD; (c) BEA-SSD Arch 2}
\label{fig:SSD}
\end{figure}

This section explores the potential of splitting the architecture during the creation of BEA (Backend for Analytics) to further minimize overhead. We specifically examine the possibility of splitting the architecture at a later stage, beyond the backbone, to not only reduce computational overhead but also maintain similar levels of uncertainty estimation and calibration.
The BEA-SSD results shown in Table ~\ref{tab:bea_results_table} by duplicating the entire detector after VGG backbone as shown in Figure ~\ref{fig:SSD}-b. The BEA-SSD has 20.1\% more parameters than the Base-SSD. The Figure ~\ref{fig:SSD}-c is optimised version of BEA and Table ~\ref{tab:overhead} shows that BEA-SSD Arch2 has just 7.8\% more parameters than the Base-SSD. By employing the same loss functions and ratios, the performance of BEA-SSD Arch2 experiences a noticeable decline. However, it is possible to restore the performance to the original BEA-SSD levels by effectively controlling the ratios of $\mathcal{L_{\mathbf{ta}}}$ and $\mathcal{L_{\mathbf{tq}}}$, as illustrated in Figure~\ref{fig:SSD-overhead}.

\begin{figure}
\centering
\begin{tabular}{cc}
\bmvaHangBox{\fbox{\includegraphics[width=5.9cm, height=2cm]{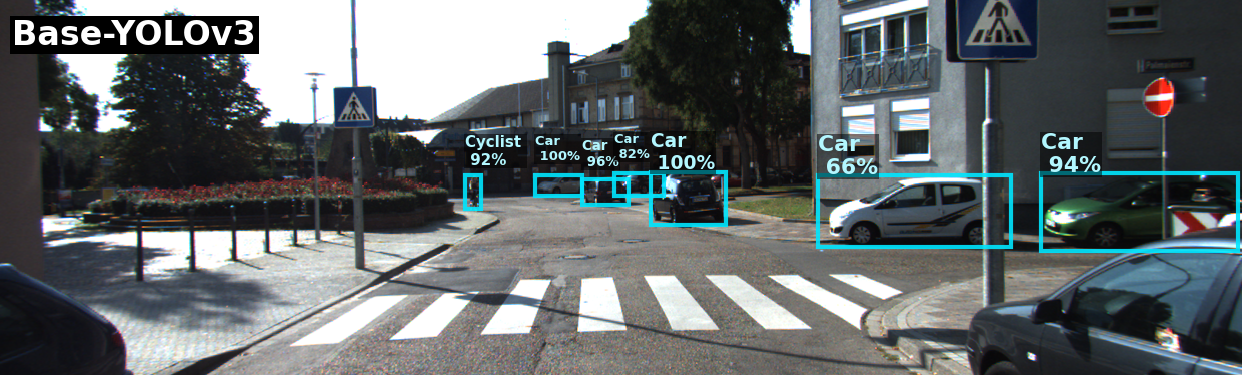}}}&
\bmvaHangBox{\fbox{\includegraphics[width=5.9cm, height=2cm]{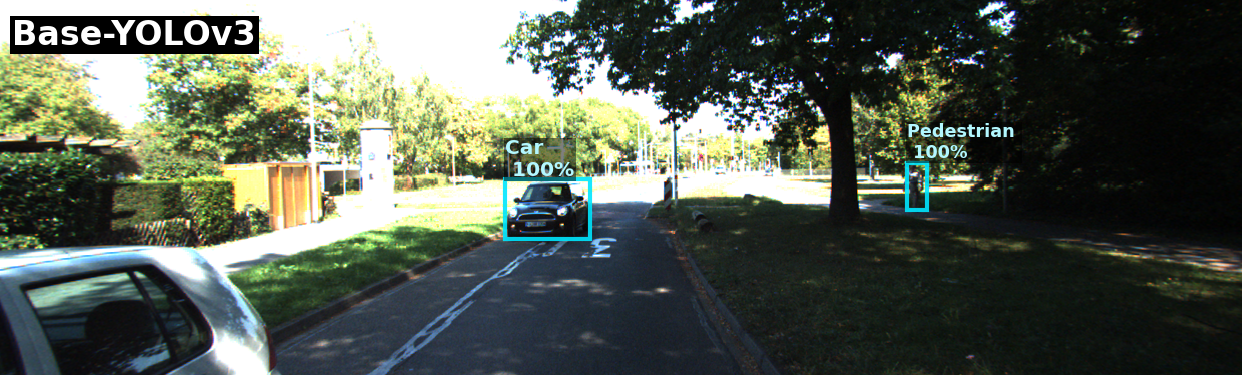}}}\\

\bmvaHangBox{\fbox{\includegraphics[width=5.9cm, height=2cm]{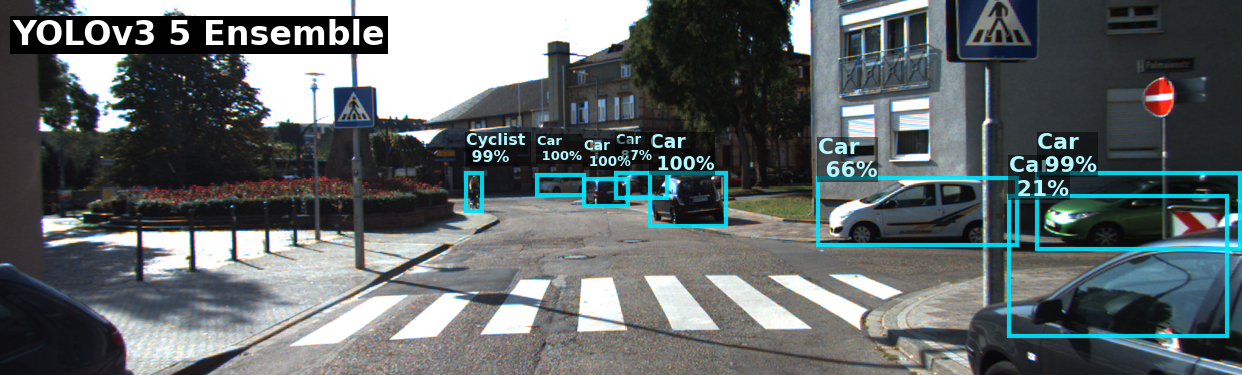}}}&
\bmvaHangBox{\fbox{\includegraphics[width=5.9cm, height=2cm]{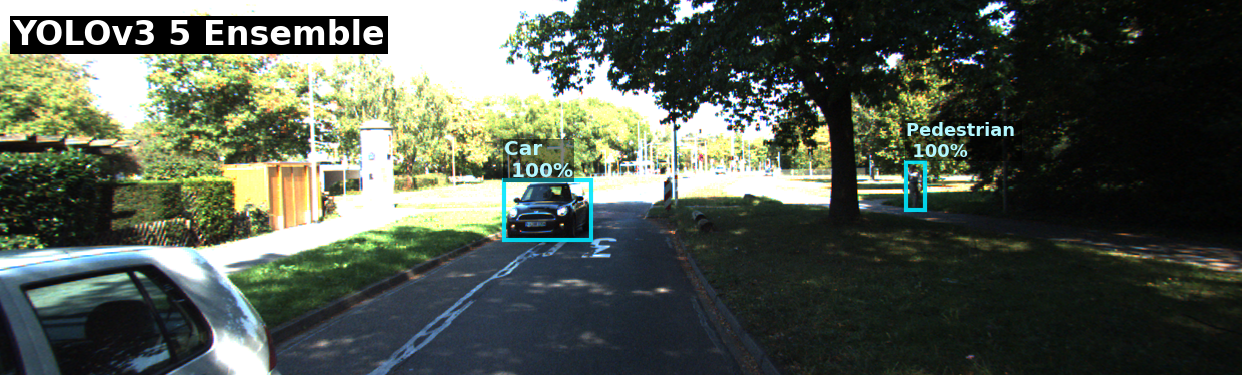}}}\\

\bmvaHangBox{\fbox{\includegraphics[width=5.9cm, height=2cm]{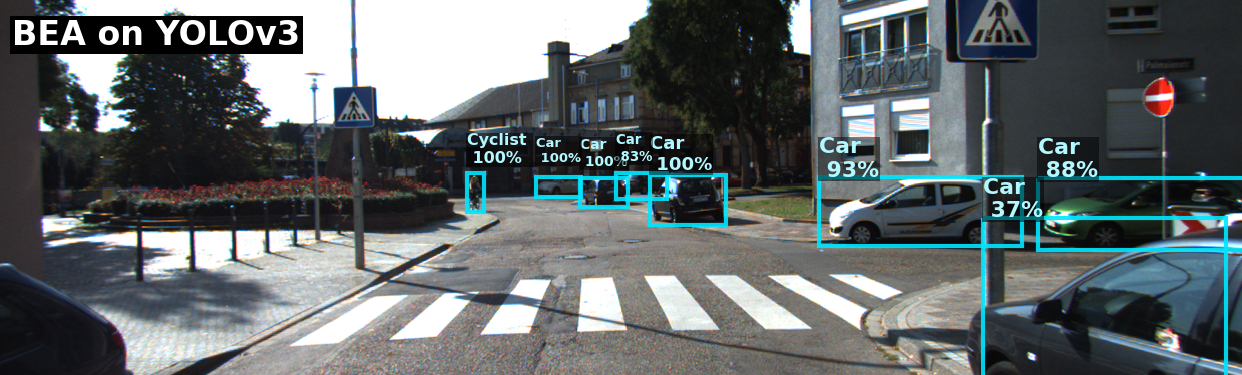}}}&
\bmvaHangBox{\fbox{\includegraphics[width=5.9cm, height=2cm]{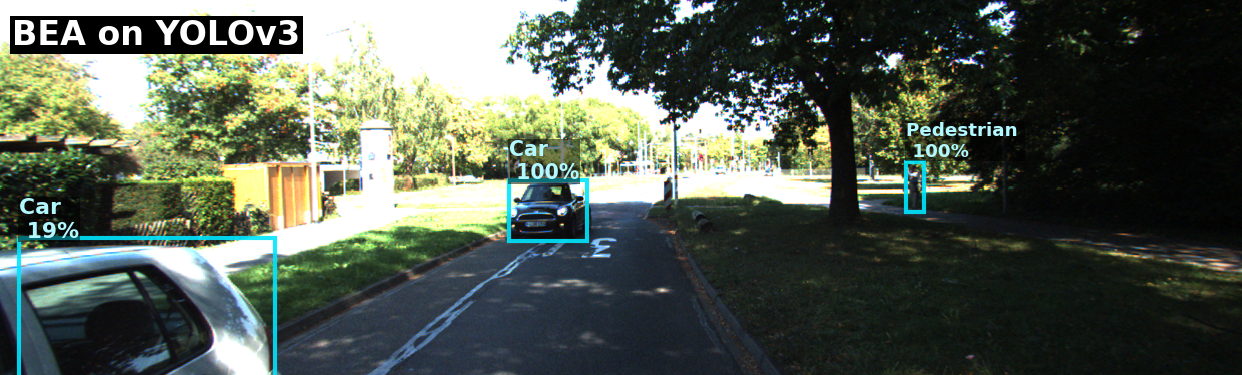}}}\\

\end{tabular}
\caption{Visual examples of BEA on YOLOv3}
\label{fig:visual_bea}
\end{figure}

\begin{figure}
\centering
\begin{tabular}{cc}
\bmvaHangBox{\fbox{\includegraphics[width=5.9cm, height=2cm]{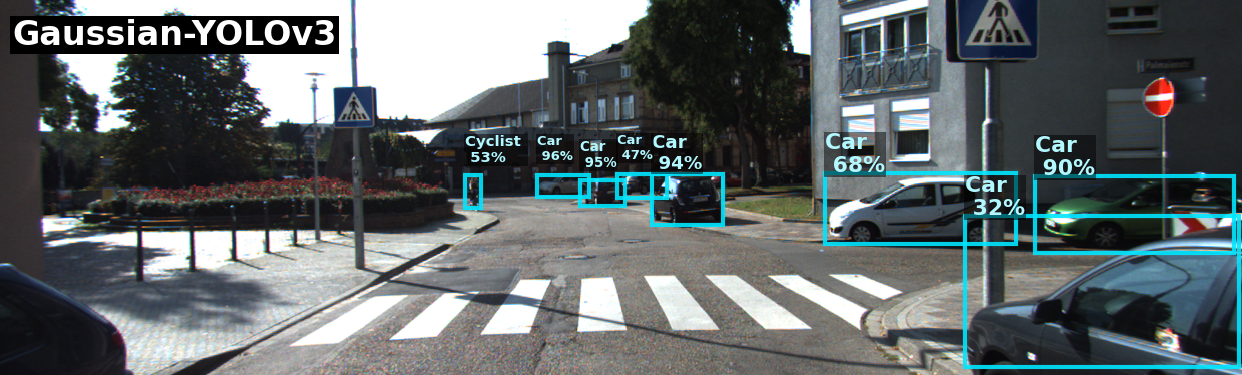}}}&
\bmvaHangBox{\fbox{\includegraphics[width=5.9cm, height=2cm]{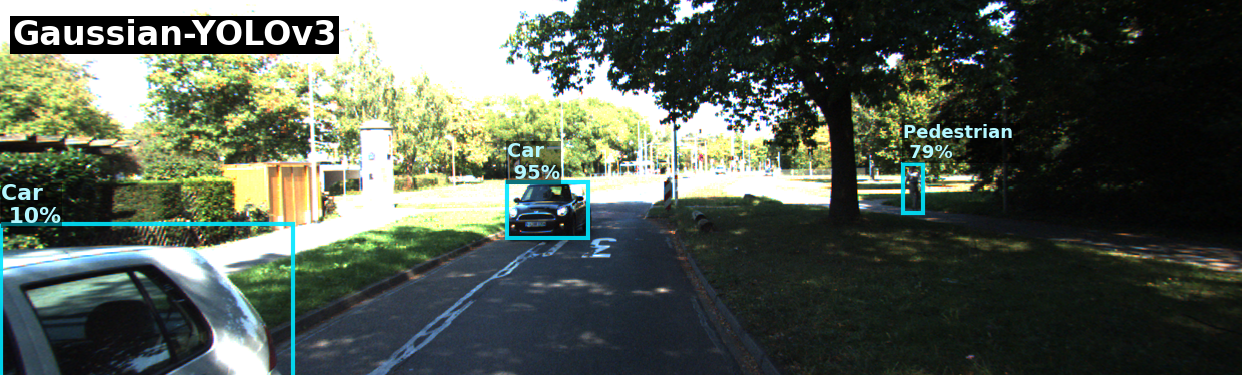}}}\\

\bmvaHangBox{\fbox{\includegraphics[width=5.9cm, height=2cm]{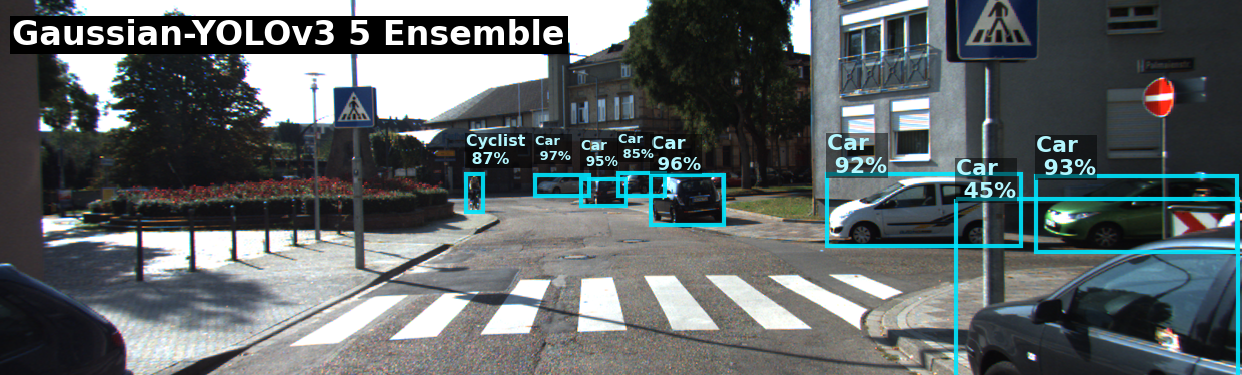}}}&
\bmvaHangBox{\fbox{\includegraphics[width=5.9cm, height=2cm]{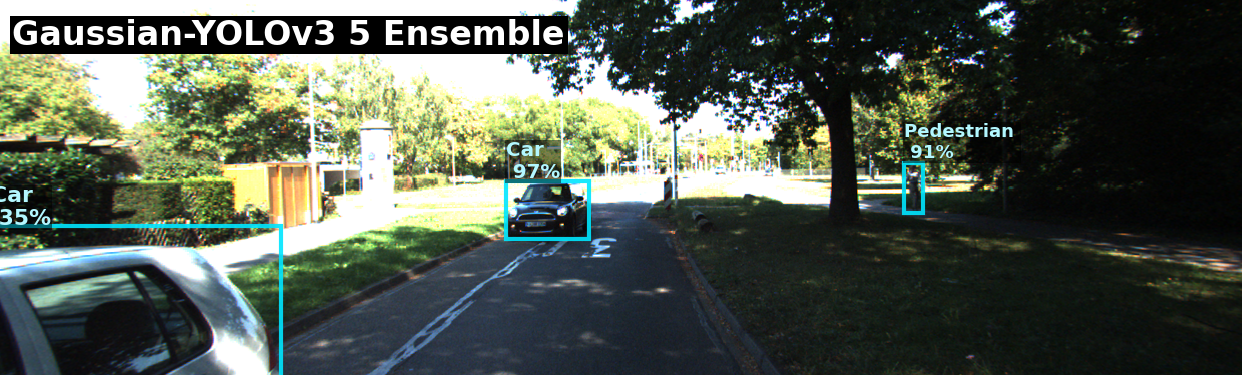}}}\\

\bmvaHangBox{\fbox{\includegraphics[width=5.9cm, height=2cm]{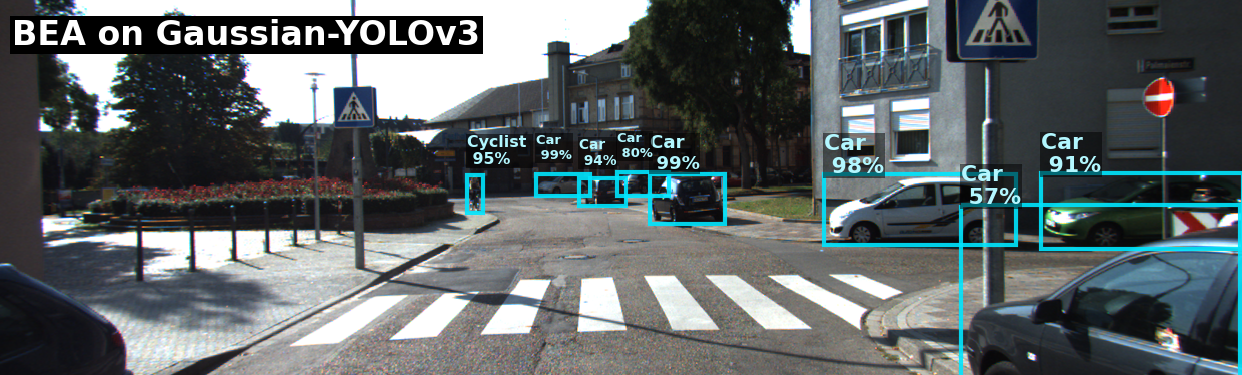}}}&
\bmvaHangBox{\fbox{\includegraphics[width=5.9cm, height=2cm]{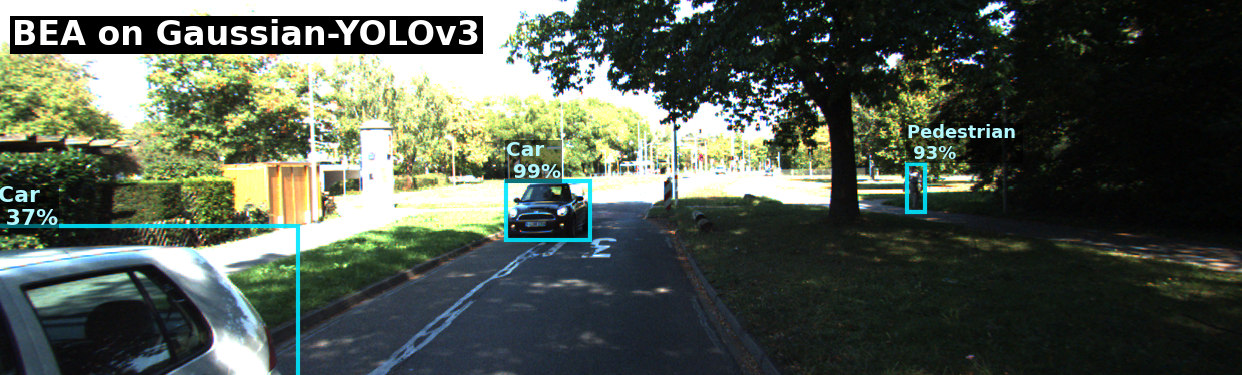}}}\\

\end{tabular}
\caption{Visual examples of BEA using Gaussian-YOLOv3}
\label{fig:visual_gaus_bea}
\end{figure}

\begin{figure*}[!ht]
    \centering
        \centering
        \includegraphics[width=0.49\textwidth, scale=0.5]{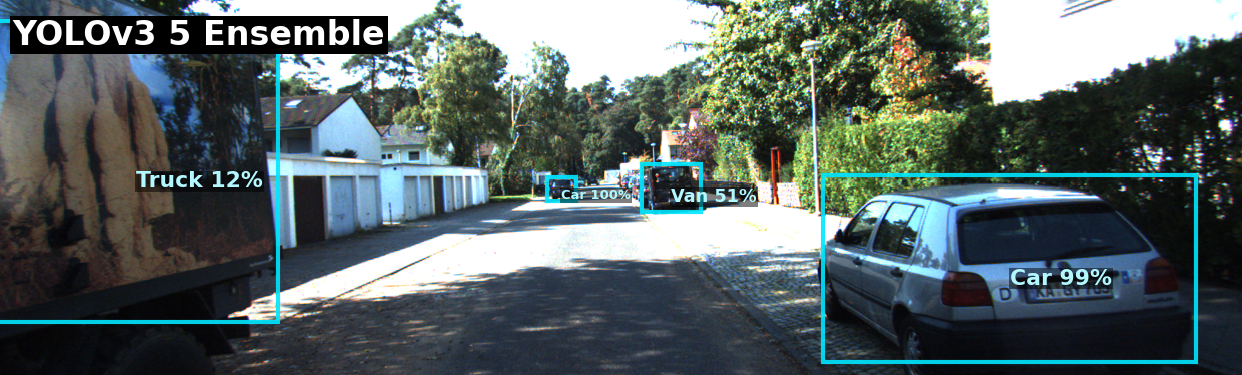}
        \includegraphics[width=0.49\textwidth, scale=0.5]{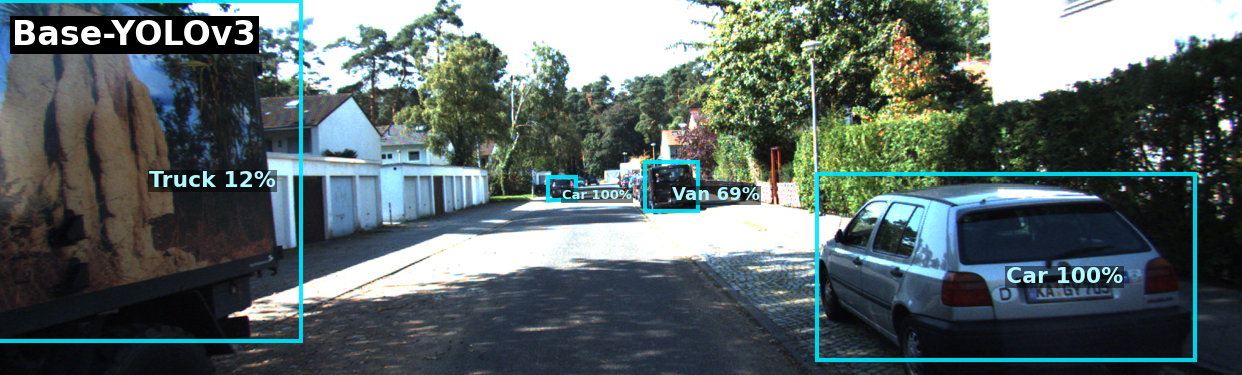}
        \includegraphics[width=0.7\textwidth]
        {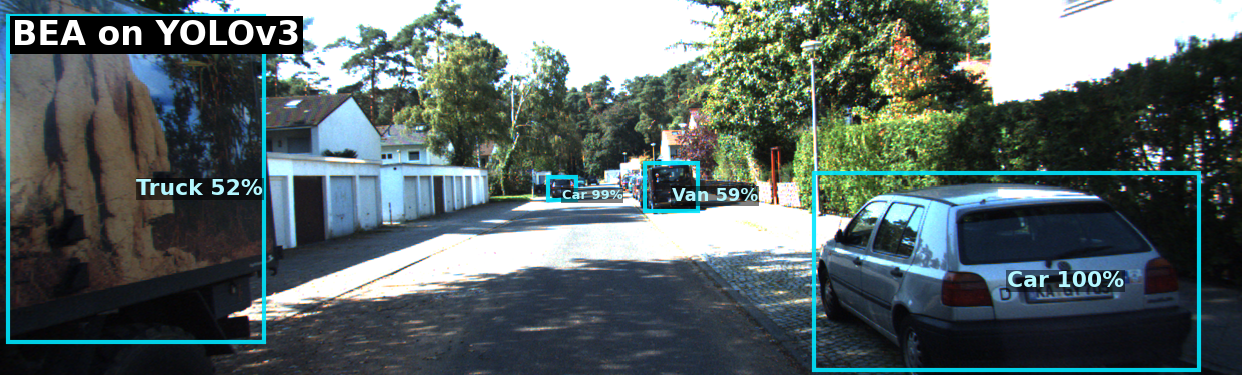}
\caption{Exotic visual example of BEA on YOLOv3: The truck with a nature poster lowers the confidence score of the prediction in non BEA models.}
\label{fig:visual_exotic_bea}
\end{figure*}

\subsection{Visual Inspection of Budding-Ensemble Architecture's performance: Figure  ~\ref{fig:visual_bea}, ~\ref{fig:visual_gaus_bea}, ~\ref{fig:visual_exotic_bea}}

\end{document}